\documentclass[journal=eds]{CUP-JNL-DTM}%

\DefineBibliographyStrings{english}{andothers={et~al.}}
\DeclareDelimFormat[textcite]{nameyeardelim}{\space}
\DeclareFieldFormat{titlecase}{%
	\ifdefstring{\currentfield}{title}%
	{\ifentrytype{book}{#1}{\MakeSentenceCase*{#1}}}%
	{#1}}
	
\usepackage{graphicx}
\usepackage{multicol,multirow}
\usepackage{amsmath,amssymb,amsfonts}
\usepackage{mathrsfs}
\usepackage{amsthm}
\usepackage{rotating}
\usepackage{appendix}
\usepackage{ifpdf}
\usepackage[T1]{fontenc}
\usepackage{newtxtext}
\usepackage{newtxmath}
\usepackage{textcomp}
\usepackage{xcolor}
\usepackage{lipsum}
\usepackage[colorlinks,allcolors=blue]{hyperref}

\usepackage{siunitx}
\usepackage{lineno}
\usepackage{longtable}
\usepackage{makecell}
\usepackage{float}
\usepackage{chngcntr}

\theoremstyle{definition}

\numberwithin{equation}{section}

\jname{Environmental Data Science}
\articletype{METHODS PAPER}
\jyear{2026}

\begin{document}

\begin{Frontmatter}

\title[Article Title]{SONATI: Spatio-temporal ocean neighbourhood asymmetric transformer for imputation of sparse, tabular data with uncertainty}

\author*[1,2]{Yvonne Jenniges}\orcid{0009-0007-7933-8002}
\author*[2,3]{Boris Peter Koch}\orcid{0000-0002-8453-731X}
\author[4,5]{Maike Sonnewald}\orcid{0000-0002-5136-9604}
\author[1]{Sebastian Maneth}\orcid{0000-0001-8667-5436}

\address[1]{\orgdiv{Databases}, \orgname{University of Bremen}, \orgaddress{\street{Bibliothekstra\ss e 1}, \city{Bremen} \postcode{28359}, \state{Bremen}, \country{Germany}}}

\address[2]{\orgdiv{Ecological Chemistry}, \orgname{Alfred-Wegener-Institut, Helmholtz-Zentrum f\"ur Polar- und Meeresforschung}, \orgaddress{\street{Am Handelshafen 12}, \city{Bremerhaven} \postcode{27570}, \state{Bremen}, \country{Germany}}}

\address[3]{\orgname{University of Applied Sciences}, \orgaddress{\street{An der Karlstadt 8}, \city{Bremerhaven} \postcode{27568}, \state{Bremen}, \country{Germany}}}

\address[4]{\orgdiv{Department of Computer Science}, \orgname{University of California}, \orgaddress{\street{1 Shields Avenue}, \city{Davis} \postcode{95616}, \state{CA}, \country{USA}}}

\address[5]{\orgdiv{School of Oceanography}, \orgname{University of Washington}, \orgaddress{\street{1410 NE Campus Parkway}, \city{Seattle} \postcode{98195}, \state{WA}, \country{USA}}}

\address[]{\textbf{Corresponding authors:} Yvonne Jenniges (\href{mailto:yvo_jen@uni-bremen.de}{yvo\_jen@uni-bremen.de}) and Boris Peter Koch (\href{mailto:boris.koch@awi.de}{boris.koch@awi.de})}

\authormark{Jenniges et al.}


\abstract{
Missing value imputation is often a prerequisite for downstream analyses. In oceanography, observational data are frequently structured in a tabular manner, highly sparse and exhibit spatio-temporal structures that classical imputers such as k-nearest neighbours or MissForest are not suited to exploit. 
We present the Spatio-temporal Ocean Neighbourhood Asymmetric Transformer for Imputation (SONATI), a masked, factorised time-space transformer operating on pre-computed, local time and space neighbourhoods, which reduces computational complexity compared to global attention. 
SONATI is designed to work with tabular data and is thus in principle able to work on both gridded and unstructured data, though evaluation here was performed on gridded North Atlantic data only. 
It provides aleatoric and epistemic uncertainties through a negative log-likelihood loss and multiple forward passes with Monte Carlo dropout. 
Systematic ablations on attention type and masking strategies guided the final architecture design.  
The results showed that SONATI outperformed classical and transformer-based methods by at least \SI{12.5}{\percent} in terms of root mean squared error (RMSE) on a large observational oceanographic dataset with key biogeochemical and physical parameters in the North Atlantic (potential temperature, salinity, oxygen, nitrate, silicate and phosphate).
Attention type and masking strategy strongly affected imputation accuracy and smoothness of the produced oceanographic fields. Best results were achieved by factorised time-space attention and combined geographic and per-sample masking. The combined masking strategy reduced the imputation error by \SI{14.3}{\percent} compared to per-sample masking only. 
Uncertainties were moderately well calibrated (miscalibration area at observed locations $0.074\pm0.012$) and moderately correlated with the imputation error.}

\end{Frontmatter}


\section*{Impact Statement}
Ocean measurements are sparse and unevenly distributed in space and time, leaving major gaps in the data. SONATI is a new deep learning method that fills these gaps by exploiting the full four-dimensional ocean structure, i.e. latitude, longitude, depth and time, where most models focus on latitude and longitude and/or time. Moreover, it reports how confident it is in every data point, distinguishing between uncertainty caused by noise and by limited knowledge. This combination of accuracy and quantified uncertainty supports follow-up analyses and informed decision making in ocean and climate research.

\section{Introduction}
Missing values are a persistent challenge in environmental and oceanographic data evaluation and applications. In ocean biogeochemistry, high missingness arises mainly from non-continuous spatio-temporal sampling (such as ship-based campaigns) but also from aspects such as sensor failures or weather conditions (e.g. cloud cover in the case of remote sensing). The resulting incomplete parameter fields hinder downstream tasks that require fully specified data points, such as clustering and forecasting. Consequently, accurate methods for imputing missing values, i.e. for filling gaps in observational data, are highly relevant for environmental downstream tasks. 

In many applications, reliable quantitative uncertainty estimates of the imputed values are important to support downstream decision-making and error propagation analyses. Uncertainty is often categorised into epistemic and aleatoric components~\citep{kendall2017, hullermeier2021}. 
This helps to identify the source of unreliable predictions and to inform whether additional data or model improvements may increase prediction reliability~\citep{haas2026}. 
Epistemic uncertainty describes the lack of knowledge of a model and can be quantified by generating multiple stochastic imputations, for example using Bayesian approaches such as Monte Carlo dropout~\citep{gal2016} or deep ensembles~\citep{lakshminarayanan2017}. 
It has proven useful for applications such as active learning, where samples with high epistemic uncertainty are preferentially selected for annotation~\citep{nguyen2022} and out-of-distribution detection, where high epistemic uncertainty may indicate previously unseen data~\citep{lu2025}. 
Aleatoric uncertainty captures noise and variability inherent to the data, increasing prediction difficulty and not reducible by additional data alone~\citep{kendall2017}. It can be estimated e.g. with a negative log-likelihood (NLL) loss, enabling the model to learn both the prediction and an input-dependent variance that down-weights noisy observations~\citep{kendall2017, barth2022}. Aleatoric uncertainty estimates help assess whether sufficiently reliable predictions can be expected despite the inherent variability of the data, thereby supporting robust decision-making~\citep{haas2026}.

Spatio-temporal imputation has been extensively researched across various applications, including transportation, economics, energy and environmental monitoring. A recent systematic review~\citep{zahmatkesh2026} categorises the field into statistical and machine learning (ML) approaches, the latter further subdivided into single and multiple imputation as well as classical ML and deep learning methods. 
Simple baselines, including fixed-value filling, k-nearest neighbours imputation (KNN,~\cite{troyanskaya2001}) and MissForest~\citep{stekhoven2012} have no integrated mechanism to model spatial or temporal structure. Such information must be explicitly encoded as input features requiring careful pre-processing. 
Geostatistical approaches such as Kriging, which weights nearby observations by spatial correlations, have been extended from modelling spatial covariance alone to also include the temporal dimension~\citep{zahmatkesh2026}. However, classical formulations rely on stationarity assumptions~\citep{zahmatkesh2026} that are frequently violated in oceanographic systems characterised by fronts, eddies and seasonal dynamics. 

In oceanographic and geophysical applications, dedicated imputation methods have been developed to exploit environmental field structures.
DINEOF (Data INterpolating Empirical Orthogonal Functions,~\cite{beckers2003}) imputes incomplete fields by iteratively computing dominant empirical orthogonal function (EOF) modes over masked observations and has been applied, for example, to fill gaps in sea surface temperature, salinity and ocean colour~\citep{beckers2006, alvera2016, alvera2025}. DINEOF provides error maps estimating the reconstruction error, though it does not distinguish between aleatoric and epistemic uncertainty.  
DINCAE 2.0 (Data-Interpolating Convolutional Auto-Encoder,~\cite{barth2022}) implements a convolutional autoencoder that can take structured or unstructured input, which is internally discretised, and produces per-pixel uncertainty estimates, representing an important step towards probabilistic oceanographic imputation. 
\citet{kadow2020} filled gaps in gridded sea and land surface temperature observations with a partial-convolution image inpainting network trained on climate model simulations. 
However, these methods have been primarily designed and evaluated for gridded surface fields and their application to sparse depth-resolved, irregularly sampled, multivariate in situ observations remains limited, as they operate on a fixed spatial grid and lack native support for depth-resolved sampling.

The broader ML literature has delivered increasingly sophisticated architectures for spatio-temporal imputation. 
Models like SAITS~\citep{du2023}, BRITS~\citep{cao2018}, SPIN~\citep{marisca2022}, ImputeFormer~\citep{nie2024} and GRIN~\citep{cini2022} exploit various architectures like attention, graph and recurrent networks. 
A common assumption is that the input data present a time series per geographic location, but in datasets such as the one from the \textit{Our Common Future Ocean in the Earth System} project (COMFORT, \cite{comfort_dataset2022}), where locations are sampled highly irregularly, the information density per time sequence is especially low. Moreover, most of these methods have been developed and benchmarked on traffic and air quality datasets; most architectures assume either a fixed, pre-defined sensor network or regularly gridded input and uncertainty quantification remains limited~\citep{zahmatkesh2026}, with notable exceptions such as the Conditional Score-based Diffusion model for Imputation (CSDI, \cite{tashiro2021}).
Tabular neural, foundation and generative imputers such as ReMasker~\citep{du2024}, TabImpute~\citep{feitelberg2026}, CFMI~\citep{simkus2025}, RefiDiff~\citep{ahamed2026}, CACTI~\citep{gorla2025} and TabINR~\citep{ochs2025} do not assume time sequence structure and have achieved strong performance on general tabular datasets. However, these methods do not natively model and exploit spatio-temporal structures. 

Here we present the Spatio-temporal Ocean Neighbourhood Asymmetric Transformer for Imputation (SONATI), a transformer-based architecture for spatio-temporal imputation of sparse oceanographic observational data. Rather than assuming a fixed sensor network or treating samples as independent tabular observations, SONATI constructs separate temporal and spatial neighbourhood graphs directly from coordinate information. This makes the architecture applicable in principle to both gridded and irregular observational datasets; empirical evaluation here is restricted to the gridded North Atlantic dataset. A time-space factorised transformer architecture processes neighbourhoods sequentially to model temporal and spatial dependencies efficiently. To quantify uncertainty, the model combines a heteroscedastic negative log-likelihood (NLL) objective with mean-variance prediction heads and Monte Carlo dropout, enabling decomposition into aleatoric and epistemic uncertainty components.

The proposed framework is evaluated on sparse multivariate oceanographic observations to address the following research questions:

\begin{enumerate} 
\item \textbf{Model performance:} How do transformer-based imputers, including ReMasker and the novel SONATI method, compare with classical baselines such as simple imputation, KNN and MissForest on sparse observational oceanographic data? 
\item \textbf{Attention mechanisms:} Which forms of attention, including self-attention, cross-attention and factorised space-time attention variants, and token processing strategies enable effective spatio-temporal imputation? 
\item \textbf{Masking strategies:} How do different masking strategies, including per-row, per-column, cylindrical and transect masking, affect model performance? 
\item \textbf{Uncertainty estimation:} How well calibrated are the uncertainty estimates produced by the newly developed SONATI and how strongly do its aleatoric and epistemic uncertainties correlate with squared imputation errors? 
\end{enumerate} 

To answer these questions, all models were systematically tuned and evaluated using root mean squared error (RMSE) and comparisons against simulation data from the Coupled Model Intercomparison Project Phase 6 (CMIP6). Uncertainty estimates were assessed using calibration curves and correlations between predicted uncertainty and squared imputation error. The proposed framework aims to improve environmental imputation tasks by providing both accurate imputations and reliable uncertainty estimates for downstream oceanographic applications.

To facilitate reuse, SONATI is released as an open-source Python PyPI package (\texttt{sonati}). 
The resulting imputed North Atlantic dataset, taking 20-year time steps from 1772 to 2020, is openly available on Zenodo~\citep{jenniges2026_sonati_data}; see the \hyperref[sec:data_availability]{Data Availability Statement} for details.


\section{Methods}
\subsection{Data}
A subset of the COMFORT dataset~\citep{comfort_dataset2022} was used focussing on the North Atlantic (\SI{0}{} -- \SI{70}{\degree N}, \SI{77}{\degree W} -- \SI{30}{\degree E}) with observational values for temperature and salinity, and for oxygen, nitrate, silicate and phosphate concentration acquired between 1772 and 2020. Data preprocessing largely followed~\citet{jenniges2025}, including spatial aggregations (\SI{1}{} $\times$ \SI{1}{\degree} horizontally, 12 depth bins with edges: [0, 50, 100, 200, 300, 400, 500, 1,000, 1,500, 2,000, 3,000, 4,000, 5,000]) and conversion to potential temperature, which makes temperatures comparable across depths by removing the effect of pressure, with the addition of a temporal dimension. 

The time resolution was set to 20 years, resulting in 13 time steps, with the last time bin being shorter ranging from 2012 to 2020. The optimal trade-off between time resolution and missingness (Figure~A1) was achieved with a time step of 11 years, which aligns with previous long-term analyses that evaluate decadal time scales~\citep{ipcc2023}. To enable a larger number of experiments and to keep computations feasible, the time step was set to 20 years. The resulting dataset consists of 637,390 grid cells and can be reproduced via example nine of the Python library \texttt{comfort-db} (version 0.1.0, \cite{jenniges2026}). 

\begin{figure}[t]%
	\FIG{\includegraphics[width=0.9\textwidth]{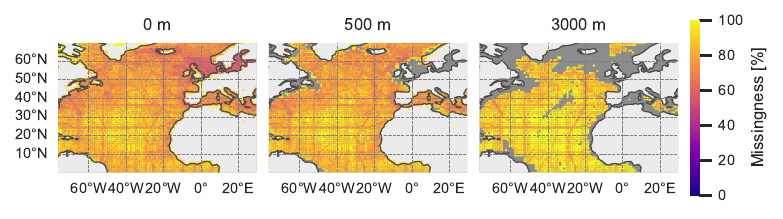}}
	{
		\caption{Data missingness, i.e. the percentage of the six parameters missing in a grid cell, averaged over time steps at selected depth levels. Grey indicates ocean cells without data and light grey shows land}
		\label{fig:data_missingness}
	}
\end{figure}

Missingness of the final dataset is highly heterogeneous across space, time and features. 
Per-parameter missingness, i.e. the percentage of grid cells with no measurement of the parameter, ranges from \SI{75}{\percent} (salinity) to \SI{91}{\percent} (nitrate).
Missingness per grid cell, referred to as missingness in the following, is defined as the percentage of the six parameters missing in a grid cell.
Despite the dataset starting in 1772, most data were captured in the time bins starting in 1972 and 1952 (Figure~A2). Early years were deliberately kept to explore extrapolation behaviour of the imputation models. 
Spatially, large, consecutive gaps remain, e.g. in the central North Atlantic (Figure~\ref{fig:data_missingness}) with the surface being sampled most (\SI{79}{\percent} missing) and coverage generally decreasing with depth (up to \SI{92}{\percent} missing), with slightly better sampling between 400 and \SI{500}{\meter}. 
The given data were primarily sampled by ships and Argo floats and hence their availability depends mainly on spatio-temporal location and sampling strategy. This is reflected in missingness patterns like ship-based transects (straight-line artefacts in Figure~\ref{fig:data_missingness}) and dominance of summer values. 
Therefore, the missingness mechanism~\citep{rubin1976} cannot be considered \textit{missing completely at random} (MCAR, missingness independent of the data), but is largely structured and consistent with a \textit{missing at random} (MAR, missingness depends only on observed features) pattern as it depends on location and time, i.e. features that are always observed. \textit{Missing not at random} (MNAR, missingness depends on unobserved values) components cannot be excluded, e.g. due to sensor limitations under specific regimes or due to avoidance of extreme or unfavourable environmental conditions.

\subsection{Data preparation}
Cross-validation was applied to estimate model generalisation abilities and to reduce overfitting. When observations are spatially or temporally autocorrelated, random splitting can overestimate predictive performance. Therefore, holdout schemes must be selected to reflect the intended application and the data’s dependence structures (\cite{roberts2017, stock2025}, Supplementary Material~B). Here, two holdout schemes were used: (i) Random holdout to assess local interpolation performance under maximal spatio-temporal leakage and (ii) block-wise holdout to test robustness to large, consecutive spatial gaps. 

In total, the data were split into six sets: Five train-validation sets used to tune model hyperparameters and one independent test set for final evaluation. Two splits were created by randomly removing \SI{10}{\percent} and \SI{20}{\percent} of observed data points. 
Three splits held out geographically consecutive blocks using a fixed time span of 20 years, three different latitude/longitude extents (5, 15 and \SI{40}{\degree}) and three depth zones ([0, 200), [200, 1,000) and [1,000, max)). 
The boundaries at 200 and 1,000 m were motivated by the conventional transitions between the epipelagic, mesopelagic and bathypelagic zones, respectively~\citep{sutton2013}. Blocks were randomly selected and assigned to the validation set until the target validation fraction was reached, while preserving the per-variable coverage within \SI{\pm 5}{\percent} of the full data set. The independent test set contained \SI{15}{\percent} of the entire data and was generated using blocks of \SI{15}{\degree} and a temporal extent of 20 years.

To prevent information leakage, validation and test sets were transformed using the scalers fitted exclusively on the respective training data. Latitude $\phi$ and longitude $\lambda$ were mapped to Cartesian coordinates on the unit sphere to encode horizontal position while avoiding discontinuities at the \SI{\pm 180}{\degree} meridian and the coordinate singularity at the poles (e.g.~\cite{russwurm2024}):

\begin{align}
	x = \cos \phi \cos \lambda \nonumber \\
	y = \cos \phi \sin \lambda \nonumber \\
	z = \sin \phi
\end{align}

The three transformed coordinates, depth (metres), time (year) as well as the six physical and biogeochemical parameters were initially mapped to $[0,1]$ using min-max scaling (MinMaxScaler, Python library \texttt{scikit-learn}, version 1.7.2,~\cite{scikit-learn}).

\subsection{Imputation methods}
Five imputation methods were tested and compared with the proposed SONATI.

The \textbf{simple imputer} computes a simple statistic across the dataset, like the per-parameter mean, to fill missing values (Python library \texttt{scikit-learn}, version 1.7.2). It does not preserve covariance structures and hence serves as a simple baseline. 

\textbf{K-nearest neighbours (KNN)} is a standard imputation approach and searches for the $k$ nearest neighbours in feature space, i.e. here across scaled spatio-temporal information and oceanographic features, and uses the neighbourhood’s mean as imputation value (\cite{troyanskaya2001}, Python library \texttt{scikit-learn}, version 1.7.2). 
KNN had previously been applied successfully to similar data averaged over time~\citep{jenniges2025}, though it does not provide uncertainties, which are crucial for reliability assessment and hence decision making. 
Another disadvantage is the computational complexity of $O(m^2n)$ with $m$ samples and $n$ features, i.e. it scales quadratically with the number of samples.

\textbf{MissForest} constructs random forest models iteratively for each feature, processed in ascending order by per-parameter missingness (\cite{stekhoven2012}, Python library \texttt{missingpy}, version 0.2.0). After an initial mean imputation, a random forest is built on observed entries of each feature using all other features as predictors and missing values are updated with the predictions. This process is repeated until convergence. MissForest proved to be effective in many tabular imputation scenarios, such as data with nonlinear relationships and mixed types~\citep{stekhoven2012, sun2023}.
		
A masked \textbf{multi-layer perceptron (MLP)} was implemented to compare the more complex transformer architectures against a simple neural network setup. The implemented MLP consists of a number of hidden layers, each containing a linear layer followed by rectified linear unit (ReLU) activation and dropout. Similar to SONATI, an NLL loss is used and the output is channelled through a mean and a variance head, each consisting of one linear layer. 

\textbf{ReMasker} is a masked autoencoder that treats each feature of a data point as a token~\citep{du2024}. In addition to the originally missing values, a random subset of observed values is masked during training. The masked input is augmented with positional encodings and processed by a transformer-based encoder-decoder, which returns the imputed values.

\subsection{Ablation studies}
Ablations were conducted on attention mechanism (how information should flow) and masking strategy (which values to hide from the model during training). Prior to hyperparameter fine-tuning, all ablation experiments were run with a fixed initial configuration five times with different seeds and evaluated on the validation set with \SI{15}{\degree} blocks. Each queried data point, i.e. the query point, attended to its 30 nearest neighbours. The attention mechanism used four heads of dimensionality 16, i.e. model dimension 64, across three layers with a feed-forward dimensionality of 256 per attention block and a dropout rate of 0.05. Models were trained for up to 100 epochs with a batch size of 128, a learning rate of \SI{1e-4}{} and early stopping after 10 epochs without validation improvement. 

\subsubsection{Attention mechanisms}
To investigate how information should flow between the current token, i.e. the query token, and its neighbourhood context, four attention architectures were tested, the last one in three variants:
(i) autoencoder, where neighbours are first encoded via self-attention and the query is then decoded by attending to the encoded neighbourhood, 
(ii) joint attention, where query and neighbours form a single sequence, 
(iii) pure cross attention, where the query attends to its neighbours one-directionally and 
(iv) factorised cross-attention, where separate cross-attention is applied sequentially over space, time and/or depth dimensions (space-time, space-time-depth and time-space attention).
Additionally, a feature mixer component was tested on each architecture that passes the features and optionally missingness information through a simple MLP without changing dimensionality, before passing the token to the attention modules. 

\subsubsection{Masking strategies}
\label{sec:methods:masking}
Masking strategies can strongly impact performance of the masked transformer and imitating missingness patterns or mixing strategies proved to be beneficial in other applications~\citep{gorla2025, kim2025, ye2025}. The masking ablation was conducted on a time-space transformer with a feature mixer, where missingness information was only provided to the attention mechanism.

Four masking strategies were compared: per-sample, per-feature, cylindrical and transect masking. Per-sample masking operates row-wise and hides a given number of randomly selected features from each sample. Per-feature masking is applied column-wise and hides a given proportion of values per feature. 

In oceanographic observations, data are often missing in large spatio-temporal chunks, i.e. there are large areas that have never been sampled. Additionally, research vessels often operate in transects, which appear as nearly straight lines on the map (Figure~\ref{fig:data_missingness}). Based on these patterns, cylindrical and transect masking were implemented. 
Cylindrical masking finds a random point in the current batch and draws a circle of radius $r$ on the latitude-longitude plane.
Transect masking hides data line-wise, similar to a ship transect, with either random direction or horizontal/vertical directions and line width $w$. Both strategies depend on latitude and longitude only, i.e. all depths and times with the identified coordinates are masked. Width $w$ and radius $r$ are Euclidean distances in scaled Cartesian coordinates, where latitude and longitude had been mapped to the unit sphere and subsequently scaled to $[0, 1]$.

\subsection{Model tuning, testing and hardware}
Hyperparameters were tuned using Optuna (\cite{akiba2019}, Python library \texttt{optuna}, version 4.6.0). For each model, a separate Optuna study with 100 trials was executed using the Tree-structured Parzen Estimator (TPE) to sample the next hyperparameter combination. In each trial, a single hyperparameter combination was evaluated across all five train-validation splits and the mean RMSE across splits served as Optuna minimisation target. 

The explored search spaces (Table~C1) and final hyperparameters are reported (Table~C2). Hyperparameter importances were obtained from Optuna’s built-in fANOVA implementation, which fits a random forest regressor to predict the objective values based on the hyperparameters~\citep{hutter2014}.

Subsequently, the models were trained using the optimal hyperparameter combination on the entire training data, including the validation subsets. 
Early stopping of the MLP and SONATI models monitored the loss on the test set.
Evaluation was then performed on the initially held-out test set. Training and evaluation on the test set were performed across five different seeds for non-deterministic models (MissForest, MLP, ReMasker, SONATI).

Computations were executed on the high-performance computing (HPC; albedo) system of the Alfred-Wegener-Institut, Helmholtz Zentrum f\"ur Polar- und Meeresforschung (AWI) using Linux OS. While the simple imputer, KNN and MissForest were tuned on CPU-based nodes (AMD Rome Epyc 7702; 64 cores, \SI{256}{\giga\byte} RAM), the neural networks (MLP, ReMasker and the novel SONATI) were tuned on GPU-based nodes (\SI{1}{\tera\byte} RAM, Nvidia A100; \SI{80}{\giga\byte} allocating \SI{16}{\giga\byte}, \SI{32}{\giga\byte} for ReMasker). The ablation studies were also executed on the GPU nodes, allocating \SI{16}{\giga\byte} of memory per job and each was run across five seeds. 

\subsection{Evaluation metrics}
Three evaluation metrics were used to evaluate and compare imputation performance. RMSE (\cite{thieu2020}, Python library \texttt{permetrics}, version 2.0.0) quantifies the magnitude of pointwise imputation errors and assigns greater weight to large deviations because errors are squared before averaging. It is non-negative and unbounded above, with zero indicating perfect agreement and lower values indicating better performance. 

Kling-Gupta efficiency (KGE,~\cite{gupta2009, kling2012}, Python library \texttt{permetrics}, version 2.0.0,~\cite{thieu2020}) combines the Pearson correlation coefficient $r_P$ of observations $y$ and predictions $\hat{y}$, bias ratio $\beta = \frac{\mu_{\hat{y}}}{\mu_y}$ (mean predicted / observed) and variability ratio $\gamma = \frac{\sigma_{\hat{y}}/\mu_{\hat{y}}}{\sigma_y / \mu_y}$ (coefficient of variation predicted / observed), capturing imputation accuracy but also distributional similarity: 

\begin{equation}
	\text{KGE}(y, \hat{y}) = 1 - \sqrt{(r_P(y, \hat{y}) - 1)^2 + (\beta(y, \hat{y}) - 1)^2 + (\gamma(y, \hat{y}) - 1)^2} 
\end{equation}

It ranges from $-\infty$ to 1, with 1 being the best achievable value. 
KGE was computed per parameter on the $[0,1]$-scaled values. The six per-parameter values of each run were median-aggregated to avoid dominance of single, poorly reproduced parameters, which can produce an arbitrarily small KGE due to the missing lower bound. These medians were then averaged over seeds. For constant predictions, variance is zero and thus $r_P$ is undefined and was set to zero. 

Mean absolute error (MAE) to model data from the \textit{Coupled Model Intercomparison Project Phase 6} (CMIP6) served as comparison of the imputation values to a simulated product and thus as a physical plausibility check. CMIP6 provides simulation data and thus time resolution to evaluate temporal trends. Data were downloaded from the Pangeo CMIP6 catalog~\citep{cmip2018}. The historical, monthly simulation data (1850 -- 2014) originated from the Earth System Model of the National Oceanic and Atmospheric Administration (NOAA) Geophysical Fluid Dynamics Laboratory (GFDL), from which the relevant parameters were selected (source ID: GFDL-ESM4, parameter IDs: thetao, so, o2, no3, si, po4). The first and the last two years were discarded and the remaining data averaged into 20-year bins and depth levels as used in this study. CMIP6 data were scaled for averaging across features using the same scalers that were fitted on the training data and absolute differences to the imputation experiments were computed per grid cell per time step. 

Uncertainties were aggregated across $T=50$ inference passes (see~\ref{sec:res:sonati}) and evaluated using calibration curves per feature~\citep{kuleshov2018}. 
Calibration measures how well the predicted uncertainty reflects the observed coverage of the prediction intervals. Specifically, for a given confidence level, it evaluates how often the true value falls within the corresponding prediction interval. For example, a well-calibrated \SI{95}{\percent} prediction interval should contain the true value in approximately \SI{95}{\percent} of all cases. 
If the observed coverage is lower than the nominal confidence level, the prediction intervals are too narrow and the model is overconfident. If the observed coverage is higher, the intervals are wider than necessary and the model is underconfident. 
Additionally, the miscalibration area, defined as the area between the ideal and the calibration curves, was reported. Correlations of the uncertainty with the squared error were computed to assess if higher uncertainties correspond to predictions where the model was less accurate. Correlations of the uncertainty with the percentage of missing features per grid cell were used to evaluate the relationship between sparsity and uncertainty. 
Lastly, to assess whether the Gaussian residual assumption of the heteroscedastic NLL loss was satisfied, standardised residuals (z-scores) were computed and their distribution was analysed across all samples. 
Calibration, residual and error analyses were computed on all observed grid cells, i.e. training and test grid cells, and thus represent the quality of the final imputation at observed locations. The comparison of epistemic and aleatoric uncertainty and the correlation with missingness were computed on all grid cells, including those without any observation.

\section{Results}
\subsection{SONATI}
\label{sec:res:sonati}
\begin{figure}[t]
	\FIG{\includegraphics[width=\textwidth]{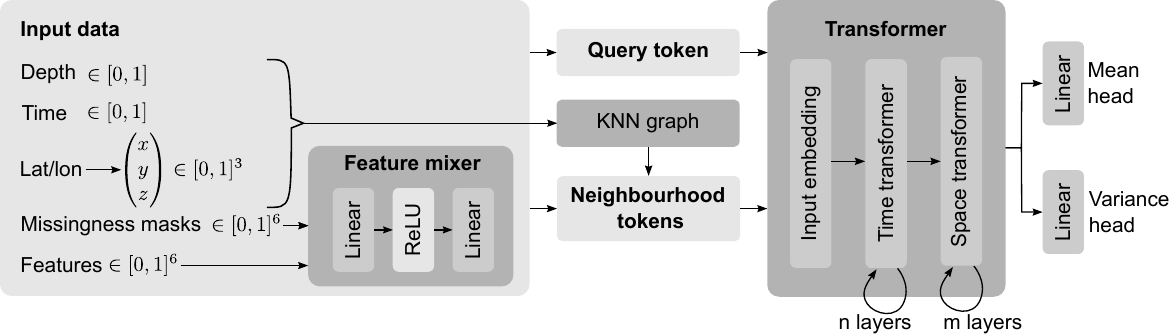}}
	{
		\caption{Architecture of SONATI, a time-space factorised masked transformer designed to impute missing input data. The input consists of a query token and its spatio-temporal neighbours. Each token contains oceanographic features and missingness masks for each feature, which are first passed through a small neural network we call a feature mixer, and spatio-temporal information. Pre-computed temporal and spatial k-nearest neighbours (KNN) graphs are then used to infer neighbours of each query point. This information is passed to the transformer, i.e. first through an embedding layer, then through $n$ layers of temporal cross attention and subsequently through $m$ layers of spatial cross attention. The cross-attention is one-directional, i.e. the query only attends to the neighbours. Finally, the mean head outputs the imputed values, the variance head outputs aleatoric uncertainties. For training, both outputs are passed to a heteroscedastic negative log-likelihood loss}
		\label{fig:sonati}
	}
\end{figure}

SONATI (Spatio-temporal Ocean Neighbourhood Asymmetric Transformer for Imputation) is a masked,  time-space factorised transformer that provides imputations along with uncertainties and operates on both gridded and ungridded data. It computes attention over temporal and spatial neighbourhoods (Figure~\ref{fig:sonati}), which is motivated by the natural spatio-temporal relationships of ocean fields.

As input, SONATI receives a query token and a list of neighbouring tokens. A token represents one datapoint in geographical space at a specific time point and contains its scaled space-time information and encoded oceanographic features and missingness masks (zero: missing, one: observed) per feature. Positional encoding is naturally provided by the spatio-temporal position, inspired by work on 3D point cloud transformers~\citep{zhao2021}. 

Separate neighbourhoods for space and time are precomputed using a k-nearest neighbours graph on all points. Using neighbourhood attention instead of global attention reduces the computational cost of the attention module from quadratic to linear in the number of input samples~\citep{hassani2023}.

The features and missingness information are first passed through a small multi-layer perceptron (MLP), the feature mixer. The token set is then linearly projected into a higher dimensional embedding space and processed by a time-space factorised transformer. Each transformer consists of a number of asymmetric cross-attention layers, i.e. they only consider the importance of neighbours for the query. The resulting latent representations are mapped to the target space by two output heads predicting the mean and variance of each oceanographic feature, following the mean-variance estimation approach~\citep{nix1994}. 
Uncertainty can be decomposed into epistemic (model variability) and aleatoric (data noise and variability) components. Epistemic uncertainty is estimated by performing multiple inference passes with Monte Carlo (MC) dropout, while the variance head, informed by a heteroscedastic NLL loss, represents aleatoric uncertainty~\citep{kendall2017}. The loss is defined as  

\begin{align}
	\mathcal{L}(y, \hat{y}, \hat{\sigma}) = \frac{1}{2D} \sum_{i} \left( \exp(-\hat{s}_i) \lVert y_i - \hat{y}_i \rVert^2 + \hat{s}_i \right)
	\label{eq:NLL_loss}
\end{align}

where $\hat{s}_i = \log \hat{\sigma}_i^2$ is the predicted log variance of a data point $i$, which is used instead of plain variance $\hat{\sigma}^2$ for numerical stability~\citep{kendall2017}, $y$ denotes the observed value, $\hat{y}$ the predicted value and $D$ is the number of samples in the current batch. 

After model training, the uncertainties were aggregated over $T$ stochastic forward passes. Epistemic uncertainty was computed as the standard deviation of the predictive means across the $T$ passes:

\begin{align}
	\hat{\sigma}_{\text{epistemic}}(x) = \sqrt{\frac{1}{T} \sum_{i=1}^{T}(\hat{\mu}_i(x) - \bar{\mu}(x))^2}
	\label{eq:epistemic}
\end{align}
	
Aleatoric uncertainty was obtained by taking the exponential of the predicted log variance $\hat{s_i}$ of each pass, averaging across $T$ passes and then taking the square root:

\begin{align}
	\hat{\sigma}_{\text{aleatoric}}(x) = \sqrt{\frac{1}{T} \sum_{i=1}^{T} \exp(\hat{s}_i(x))} = \sqrt{\frac{1}{T} \sum_{i=1}^{T} \hat{\sigma}_i^2(x)}
	\label{eq:aleatoric}
\end{align}

Total uncertainty was computed as the square root of the summed epistemic and aleatoric variances:

\begin{align}
	\hat{\sigma}_{\text{total}}(x) = \sqrt{\hat{\sigma}_{\text{epistemic}}^2(x) + \hat{\sigma}_{\text{aleatoric}}^2(x)}
	\label{eq:total_uncertainty}
\end{align}

During training, an additional random feature masking strategy was applied in each epoch for self-supervised reconstruction. The strategy combined masking of a specified proportion of features and geographic masking (per-sample and cylindrical masking, see Section~\ref{sec:methods:masking}). The loss was only computed on masked but originally observed entries, ignoring permanently missing values, indicated by the feature masks.

\subsection{Model comparison}
The evaluation of the tuned imputation models on the test set (Table~\ref{tab:model_comparison}) revealed that SONATI achieved the lowest overall RMSE ($0.0253 \pm 0.0003$) and highest KGE ($0.93 \pm 0.01$).
Its RMSE was 0.0036 (\SI{12.5}{\percent}) lower than that of ReMasker and 0.0042 (\SI{14.2}{\percent}) lower than that of MissForest. KNN showed a substantially higher RMSE (0.0461), while MLP and mean imputation performed considerably worse ($0.1322 \pm 0.0595$ and 0.1233, respectively, both with negative KGE). CMIP6 differences fell in a similar range for all models, except for mean imputation with at least twice the MAE of the other models. 

\begin{table}[htbp]
	\centering
	\caption{Imputation results of the six tuned models on the test set in $[0,1]$-scaled space. Root mean squared error (RMSE) was computed across all test values and hence weights parameters by their number of test observations. 
	Kling-Gupta efficiency (KGE) is the median across parameters per run.
	The mean absolute error (MAE) to CMIP6 simulation data was computed at originally missing positions and averaged across parameters.
	Values are reported as mean $\pm$ standard deviation across five model runs with different seeds. Mean imputation and KNN are deterministic and were run once}
	\label{tab:model_comparison}
	\begin{tabular}{llll}
		\toprule
		Model & RMSE $\downarrow$ & KGE $\uparrow$ & CMIP6 MAE $\downarrow$ \\
		\midrule
		MLP & 0.1322 $\pm$ 0.0595 & $-$1.36 $\pm$ 3.70 & 0.0228 $\pm$ 0.0008 \\
		Mean & 0.1233 & $-$0.41 & 0.0602 \\
		KNN & 0.0461 & 0.86 & 0.0300 \\
		MissForest & 0.0295 $\pm$ 0.0002 & 0.90 $\pm$ 0.01 & 0.0244 $\pm$ 0.0004 \\
		ReMasker & 0.0289 $\pm$ 0.0004 & 0.84 $\pm$ 0.01 & 0.0233 $\pm$ 0.0008 \\
		SONATI & \textbf{0.0253 $\pm$ 0.0003} & \textbf{0.93 $\pm$ 0.01} & \textbf{0.0220 $\pm$ 0.0001} \\
		\bottomrule
	\end{tabular}
\end{table}

While SONATI dominated overall, the per-parameter metrics (Table~\ref{tab:model_comparison_per_var}) showed a more nuanced ranking. SONATI achieved the lowest RMSE for five of the six parameters, with largest differences to the second-ranking model for potential temperature (0.0049 lower than ReMasker), oxygen (0.0035 lower than MissForest) and phosphate (0.0022 lower than ReMasker). Silicate was best reproduced by MissForest, although the difference to SONATI was small (0.0007). 

\begin{table}[htbp]
	\centering
	\caption{Root mean squared error (RMSE) of the six tuned models on the test set per $[0,1]$-scaled parameter. Standard deviations are reported across five model runs with different seeds}
	\label{tab:model_comparison_per_var}
	\resizebox{\textwidth}{!}{%
		\begin{tabular}{lllllll}
			\toprule
			Model & Potential temperature & Salinity & Oxygen & Nitrate & Silicate & Phosphate \\
			\midrule
			MLP & 0.2388 $\pm$ 0.1342 & 0.0776 $\pm$ 0.0139 & 0.0894 $\pm$ 0.0157 & 0.0381 $\pm$ 0.0122 & 0.0457 $\pm$ 0.0084 & 0.0301 $\pm$ 0.0095 \\
			Mean & 0.1934 & 0.0919 & 0.1417 & 0.0277 & 0.0444 & 0.0241 \\
			KNN & 0.0613 & 0.0391 & 0.0578 & 0.0141 & 0.0378 & 0.0147 \\
			MissForest & 0.0410 $\pm$ 0.0002 & 0.0133 $\pm$ 0.0002 & 0.0391 $\pm$ 0.0005 & 0.0113 $\pm$ 0.0001 & \textbf{0.0274 $\pm$ 0.0002} & 0.0168 $\pm$ 0.0009 \\
			ReMasker & 0.0366 $\pm$ 0.0002 & 0.0137 $\pm$ 0.0003 & 0.0410 $\pm$ 0.0011 & 0.0131 $\pm$ 0.0002 & 0.0307 $\pm$ 0.0005 & 0.0141 $\pm$ 0.0004 \\
			SONATI & \textbf{0.0317 $\pm$ 0.0005} & \textbf{0.0129 $\pm$ 0.0010} & \textbf{0.0356 $\pm$ 0.0011} & \textbf{0.0100 $\pm$ 0.0002} & 0.0281 $\pm$ 0.0005 & \textbf{0.0119 $\pm$ 0.0010} \\
			\bottomrule
		\end{tabular}
	}
\end{table}

The KGE results (Table~D1) showed very small differences between the three best-performing models (SONATI, ReMasker and MissForest) for potential temperature, salinity, oxygen and silicate with a maximum difference of 0.04. For nitrate and phosphate, SONATI achieved the highest KGE ($0.91 \pm 0.02$ and $0.82 \pm 0.04$, respectively) with differences of 0.08 and 0.07 to the second-ranking MissForest. 

Similarly, differences to the CMIP6 simulation data per-parameter (Tables~D2 and D3) were mostly small. SONATI produced fields closest to CMIP6 for salinity and silicate, while ReMasker performed best for nitrate and phosphate, MissForest for potential temperature and MLP for oxygen.

\begin{figure}[t]%
	\FIG{\includegraphics[width=\textwidth]{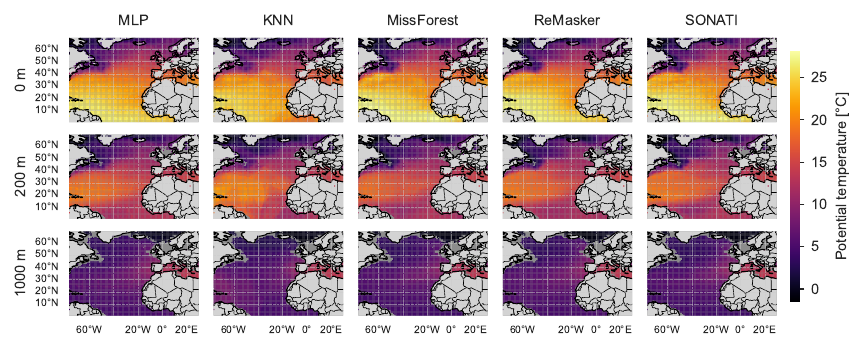}}
	{
		\caption{Imputation of potential temperature by a multi-layer perceptron (MLP), k-nearest neighbours (KNN) algorithm, MissForest, ReMasker and the novel SONATI, averaged over time and five seeds at three selected depth levels. KNN is deterministic and was hence run once. Grey indicates ocean cells without data and light grey shows land}
		\label{fig:pott_imputation}
	}
\end{figure}

\begin{figure}[t]%
	\FIG{\includegraphics[width=\textwidth]{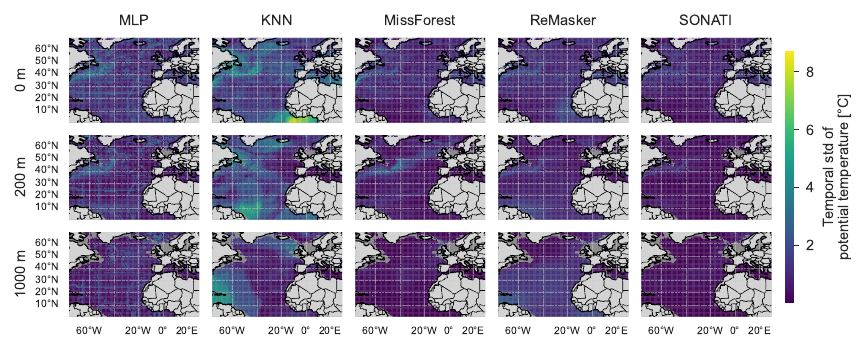}}
	{
		\caption{Standard deviation over time of the seed-averaged potential temperature imputation at three selected depth levels, generated by a multi-layer perceptron (MLP), k-nearest neighbours (KNN) algorithm, MissForest, ReMasker and the novel SONATI. KNN is deterministic and was hence run once. Grey indicates ocean cells without data and light grey shows land}
		\label{fig:pott_imputation_std}
	}
\end{figure}

Visually, the imputed temperature fields (Figure~\ref{fig:pott_imputation}) and the standard deviation of predictions over time (Figure~\ref{fig:pott_imputation_std}) of MLP exhibited ship-transect artefacts despite the plots showing an average over five seeds and for Figure~\ref{fig:pott_imputation} additionally over time. 
KNN resulted in fragmented prediction and standard deviation fields with partly sharp edges. 
SONATI, MissForest, ReMasker and MLP showed the highest temporal variability at the surface and in the Gulf Stream area, while MLP and ReMasker additionally varied at the West African coast and in deeper layers. 

\begin{figure}[t]%
	\FIG{\includegraphics[width=\textwidth]{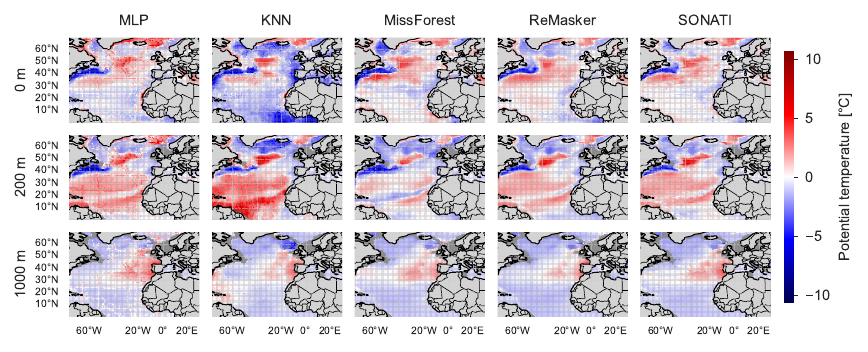}}
	{
		\caption{Mean differences (five-seed-averaged imputation $-$ CMIP6 simulation data) of potential temperature at originally missing grid cells for imputations by a multi-layer perceptron (MLP), k-nearest neighbours (KNN) algorithm, MissForest, ReMasker and the novel SONATI, averaged over time at three selected depth levels. Grey indicates ocean cells without data and light grey shows land}
		\label{fig:pott_imputation_cmip6_diff}
	}
\end{figure}

Deviations of potential temperature from the CMIP6 simulation product (Figure~\ref{fig:pott_imputation_cmip6_diff}) revealed similar patterns across all tested models: The western part of the Gulf Stream along the American coast was underestimated and the continuation from about 40 -- \SI{20}{\degree W} overestimated. Overestimation also happened in the northern area at the surface, the central area and the area off Europe at \SI{1000}{\meter} depth. 
However, the models still deviated in amplitude and locally in extent and direction. For example, KNN strongly underestimated the southern area, while the other models tended towards overestimation or lighter underestimation. 

\subsection{Ablation}
\subsubsection{Attention mechanism}
\begin{figure}[t]%
	\FIG{\includegraphics[width=0.9\textwidth]{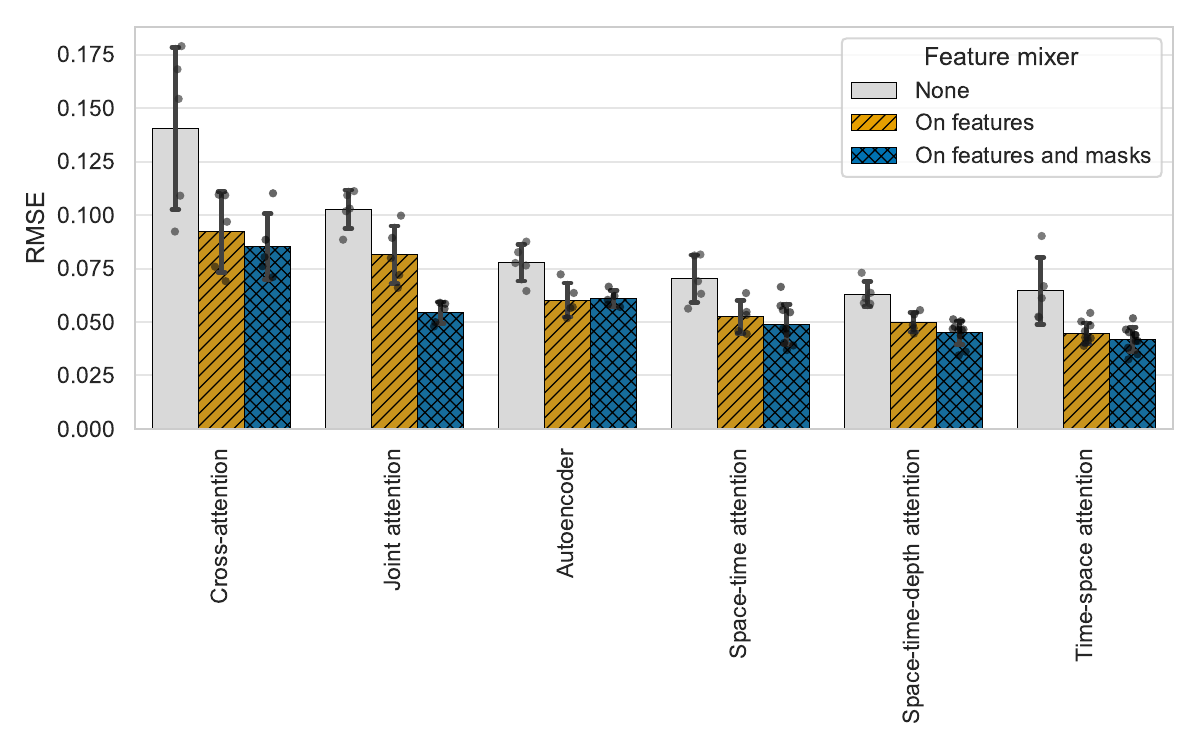}}
	{
		\caption{Comparison of attention mechanisms with and without a feature mixer, which received features as input and optionally missingness masks. The plot shows the RMSE computed across all test values, pooled over features, space and time as mean and standard deviation over five seeds}
		\label{fig:attention_mechanisms}
	}
\end{figure}

The best overall performance as quantified by RMSE, KGE and CMIP6 MAE (Figure~\ref{fig:attention_mechanisms}; Table~E1) was obtained by time-space attention with features and masks as input to a feature mixer (RMSE: $0.0419 \pm 0.0057$, KGE: $0.84 \pm 0.04$, CMIP6 MAE: $0.0232 \pm 0.0017$), closely followed by time-space attention with only features as input to a feature mixer. 
Third and fourth ranks were also occupied by factorised transformers, with space-time-depth attention and space-time attention using features and missingness masks achieving RMSEs of $0.0449 \pm 0.0055$ and $0.0489 \pm 0.0095$, respectively. 
The autoencoder, joint and cross-attention architectures yielded higher RMSEs  in their respective best configuration ($0.0602 \pm 0.0079$, $0.0543 \pm 0.0051$ and $0.0852 \pm 0.0154$, respectively).

Among models without a feature mixer, the three factorised transformer architectures also achieved the lowest RMSE values, ranging from $0.0631 \pm 0.0059$ for space-time-depth attention to $0.0702 \pm 0.0111$ for space-time attention. 

Using a feature mixer (with features as the only input) reduced the overall RMSE for all architectures. Relative to the corresponding configurations without a feature mixer, RMSE decreased by 
0.0485 (\SI{34.5}{\percent}) for cross-attention, 
0.0199 (\SI{30.8}{\percent}) for time-space attention, 
0.0178 (\SI{25.4}{\percent}) for space-time attention, 
0.0176 (\SI{22.6}{\percent}) for the autoencoder, 
0.0132 (\SI{20.9}{\percent}) for space-time-depth attention and 
0.0214 (\SI{20.8}{\percent}) for joint attention. 

The addition of missingness masks to the feature mixer input further reduced RMSE compared to the feature mixer with features only by 
0.0271 (\SI{33.3}{\percent}) for joint attention, 
0.005 (\SI{10}{\percent}) for space-time-depth attention, 
0.0069 (\SI{7.5}{\percent}) for cross-attention, 
0.0035 (\SI{6.7}{\percent}) for space-time attention, 
0.0028 (\SI{6.3}{\percent}) for time-space attention, 
whereas the autoencoder showed a marginal increase of 0.0006 (\SI{1}{\percent}).

\begin{figure}[t]%
	\FIG{\includegraphics[width=\textwidth]{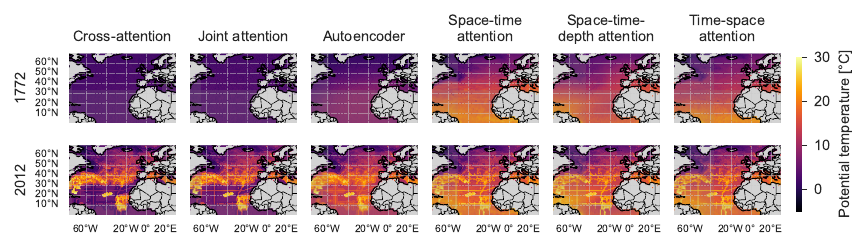}}
	{
		\caption{Imputed potential temperature fields produced by six different attention mechanisms without a feature mixer at the surface level in two selected 20-year bins (1772 and 2012), averaged across five seeds. Grey indicates ocean cells without data and light grey shows land}
		\label{fig:attention_mechanisms_pott}
	}
\end{figure}

Imputed potential temperature fields (Figure~\ref{fig:attention_mechanisms_pott}) revealed clear differences. 
In the 1772 time bin at \SI{100}{\percent} missingness, cross and joint attention produced a near-uniform field, similar to the autoencoder that was able to infer slightly elevated temperatures in the central North Atlantic. The three factorised transformers showed a stronger time signal, with space-time attention exhibiting most structure. 
In the 2012 time bin (\SI{77.6}{\percent} missing), all models constructed ship-transect artefacts and different degrees of background-foreground separation, visible as sharp edges. The latter were especially pronounced in cross- and joint attention, while the autoencoder slightly softened the edges. The factorised models' output was overall smoother, though artefacts remained.

\subsubsection{Masking strategies}
\begin{figure}[t]
	\centering
	\FIG{\includegraphics[width=\textwidth]{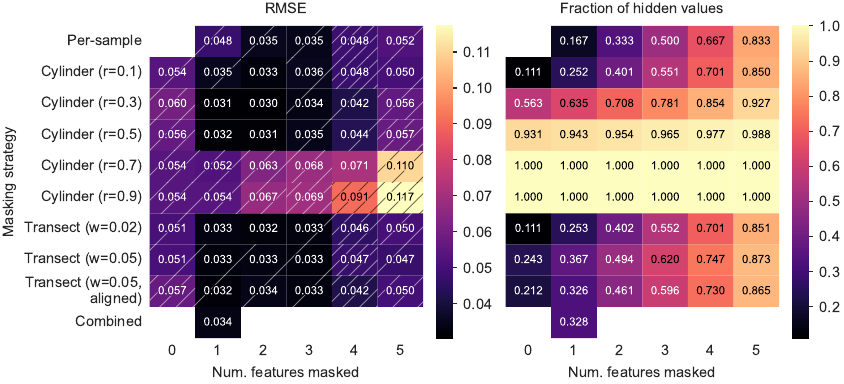}}
	\caption{RMSE of time-space attention under various masking strategies, applied alone and in combination, evaluated on the \SI{15}{\degree}-block validation set (left) and the corresponding fraction of hidden values (right). Columns denote the number of features masked per data point in addition to the row’s masking strategy. The first column corresponds to that strategy applied on its own, without extra per-sample masking. The first row shows per-sample masking alone. The middle rows combine per-sample masking with cylindrical masking  (spatial radius $r\in\{0.1, 0.3, 0.5, 0.7, 0.9\}$), simulating regional data gaps and transect masking (width $w\in\{0.02, 0.05\}$), simulating ship-track gaps, applied in a random or grid-aligned direction. The last row shows per-sample, transect and cylindrical masking simultaneously. Cell colour encodes RMSE magnitude, while hatching indicates the coefficient of variation of RMSE across the five seeds for that cell, min-max normalised within the heatmap. Denser hatching marks a less stable RMSE across seeds}
	\label{fig:masking_strategies}
\end{figure}

The smallest RMSE ($0.030 \pm 0.001$, KGE $0.89 \pm 0.02$, CMIP6 MAE $0.023 \pm 0.000$) was achieved by combining per-sample masking of two features with cylindrical masking at radius 0.3 with \SI{70.8}{\percent} of values effectively hidden (Figures~\ref{fig:masking_strategies} and E1). 
A cluster of configurations reached similarly low RMSE (0.030 -- 0.036, KGE 0.82 -- 0.90, CMIP6 MAE 0.023 -- 0.024): Cylindrical masking at radius 0.1 -- 0.5, transect masking at any tested width or orientation and the combined strategy, each paired with one to three masked features per sample. Per-sample masking alone with two or three masked features also fell into that range. 
With medium mask ratios (0.2 -- 0.7), per-feature masking achieved a comparable RMSE (0.0326 -- 0.0345, KGE 0.86 -- 0.88, CMIP6 MAE 0.0228 -- 0.0239, Table~E2).

Highest errors (RMSE 0.052 -- 0.117, KGE 0.19 -- 0.71, CMIP6 MAE 0.024 -- 0.030) were produced by combining per-sample masking with large-radius cylinders ($r \geq 0.7$), increasing mostly monotonically with the number of masked features. 
Masking four or five features per sample also produced  high errors across most strategies (RMSE 0.042 -- 0.117), corresponding to 66.7 -- \SI{100}{\percent} hidden values. 
Geographic masking alone (Figure~\ref{fig:masking_strategies}, first column) ranged from an RMSE of $0.051 \pm 0.004$ (transect with $w = 0.02$) to $0.060 \pm 0.009$ (cylinder with radius $r = 0.3$), with KGE between 0.63 and 0.72 and CMIP6 MAE between 0.023 and 0.025 at 11.1 -- \SI{100}{\percent} effectively hidden values. 


\subsection{Uncertainties}
\begin{figure}[t]%
	\FIG{\includegraphics[width=0.6\textwidth]{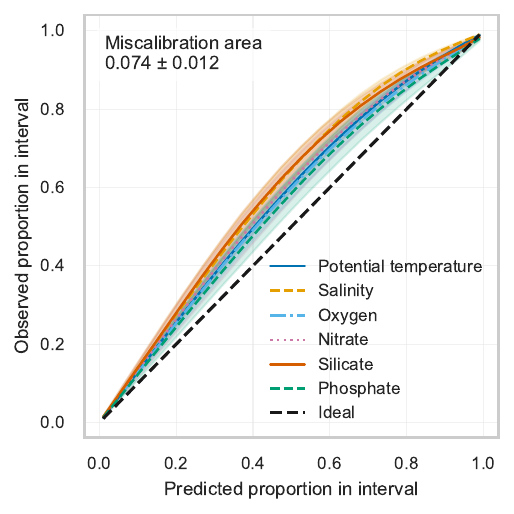}}
	{
		\caption{Uncertainty calibration per feature $\pm 1$ standard deviation across five seeds, with a dashed line indicating ideal calibration}
		\label{fig:uncertainty_calibration}
	}
\end{figure}

The total uncertainties, i.e. aleatoric and epistemic uncertainties combined, of the final SONATI model were generally close to the ideal calibration curve with a mean miscalibration area of $0.074 \pm 0.012$ across the five seeds (Figure~\ref{fig:uncertainty_calibration}). However, the model was slightly underconfident, i.e. the observed proportion in an interval exceeded the predicted proportion, especially for salinity (miscalibration area of $0.095 \pm 0.020$) and silicate (miscalibration area of $0.092 \pm 0.019$). Phosphate uncertainties (miscalibration area of $0.054 \pm 0.026$) were closest to the ideal. 

The z-score residual analysis showed that the residuals were consistent with the Gaussian assumption across features (Table~F1): Z-means for the features ranged between \SI{-0.099}{} and 0.093 and were thus close to zero and z-standard deviations ranged from 0.83 to 0.94 and were thus slightly lower but close to one.

For all features, uncertainties correlated moderately with the squared error (Spearman's $\rho=0.43 - 0.63$), with the strongest correlation for salinity and potential temperature (both 0.63) and the weakest for oxygen (Figure~F1). 
Aleatoric uncertainty consistently dominated epistemic uncertainty across all features (Figure~F2), with the median (across all grid cells and seeds) differing by a factor between about four and six. The aleatoric uncertainty distributions showed pronounced outliers, most notably for silicate, potential temperature and salinity. The highest epistemic and aleatoric uncertainties were found in potential temperature and oxygen. 
Uncertainty-missingness correlations were positive for both uncertainty types for all features (Figure~F3), with oxygen having the strongest aleatoric correlation ($\rho=0.26 \pm 0.02$) and phosphate the strongest epistemic correlation ($\rho=0.28 \pm 0.03$). 

\section{Discussion}
\subsection{Model comparison}

SONATI achieved the best overall RMSE, KGE and CMIP6 MAE. In RMSE, it was followed by ReMasker and MissForest with KNN, MLP and mean imputation performing substantially worse.
In KGE, the ranking differed (MissForest, KNN, ReMasker) with MLP and mean imputation being negative. 

The strong performance of MissForest is noteworthy due to the model's simplicity, though random-forest-based imputation is known for its capability to handle non-linear, complex interactions in tabular data effectively~\citep{stekhoven2012, golchian2026}. 
SONATI's lower RMSE overall, but also on five of the six parameters with particularly pronounced differences for potential temperature, oxygen and phosphate, suggests that explicitly modelling spatio-temporal interactions provided predictive value beyond the feature relationships captured by MissForest. Moreover, MissForest does not natively provide uncertainty estimates as opposed to SONATI, though theoretical approaches exist to integrate epistemic and aleatoric uncertainty into the algorithm (e.g.~\cite{shaker2020}).

ReMasker, which also does not natively contain uncertainty estimates, ranked second in RMSE on the test dataset, but only fourth in KGE. This discrepancy suggests that, despite relatively accurate point-wise predictions, ReMasker may reproduce the variability, correlation structure or mean behaviour of the observed data less consistently. In contrast, SONATI combines neighbourhood-based attention with cross-feature information through the feature mixer, potentially allowing it to better capture the joint structure of the oceanographic parameters. 

KNN performed worse than SONATI and produced partly polygonal structures and sharp edges in the imputed fields, likely reflecting abrupt changes in the set of nearest neighbours used for prediction. 
Although SONATI also operates on predefined local neighbourhoods, its attention mechanism learns context-dependent weights for individual neighbours rather than applying distance-based averaging, which may have contributed to the smoother transitions observed in its imputed fields, though the main contributor seemed to be the feature mixer component (Figure~E2). 
Since RMSE favours smooth predictions, these gradual transitions may also have contributed to SONATI's lower RMSE. 

Mean imputation performed second-worst in overall RMSE and KGE. 
Filling missing values with their feature mean removes spatial and temporal variability and dilutes relationships between variables. 
Consequently, dynamics such as higher nutrient values at river mouths cannot be represented, rendering mean imputation an unsuitable approach for imputing sparse oceanographic data, though providing a useful baseline for the comparison.
The KGE of mean imputation ($-0.41$) is close to $1-\sqrt{2}\approx-0.41$, the value of a constant predictor that matches the observed mean and has no variability or correlation with the observations~\citep{knoben2019}.

The MLP performed on average worse than the mean imputation baseline in both RMSE ($0.1322 \pm 0.0595$ vs. 0.1233) and KGE ($-1.36 \pm 3.70$ vs. $-0.41$), though its validation RMSE during tuning (0.030, Table~C2) was comparable to that of the other models. Its test RMSE varied strongly between seeds (0.050--0.189), suggesting unstable generalisation to the held-out test blocks, and it exhibited visual ship-transect artefacts.

The CMIP6 differences did not reliably track imputation skill since MLP attained the second-lowest overall CMIP6 MAE and the best MAE for oxygen despite its substantially worse RMSE and KGE. 
This suggests that, in this study, similarity to the smooth, simulated fields might be a less reliable indicator for imputation quality and model ranking, but rather a coarse physical plausibility check. Mean imputation showed clearly larger CMIP6 deviations than the other methods, consistent with its poor quantitative and visual performance. 
The similar spatial patterns of the CMIP6 differences across methods may partly reflect properties shared across the reconstruction and reference data, rather than imputation skill alone. Potential contributors might be biases in the CMIP6 simulations, predominant sampling during summer and the aggregation of heterogeneous observations into 20-year means. 

Several models showed increased temporal variability in the Gulf Stream region, particularly near the surface. Because similar patterns occurred in MLP, MissForest, ReMasker and SONATI, they are more likely to reflect a structure in the underlying data than an effect of a single algorithm. At \SI{200}{\meter} depth, MissForest produced the clearest continuous band of high variability along the boundary between warm subtropical and colder northern waters. Oceanographic studies commonly approximate the northern boundary of the Gulf Stream using the line at which temperature at \SI{200}{\meter} depth equals \SI{15}{\celsius} (isotherm). Movements of this boundary cause temperature changes at fixed locations and can therefore produce a band of high temporal standard deviation \citep{joyce2000,wolfe2019}. Farther east, the band follows the Gulf Stream Extension and North Atlantic Current, which are also known to vary over time \citep{breckenfelder2017}. Its weaker expression in SONATI nevertheless indicates that the estimated magnitude of this variability depends on the imputation method. 

In contrast, some structures were more likely caused by the algorithms or the sampling pattern. The straight and intersecting lines produced by MLP resembled the routes along which the original ship-based measurements were collected. KNN generated sharply separated, partly polygonal regions, which are consistent with abrupt changes in the set of neighbouring observations used for prediction. Natural ocean fronts can also produce sharp temperature gradients, but they would not normally reproduce ship routes or the geometric boundaries generated by a nearest-neighbour algorithm. The alignment of these structures with the sampling or algorithmic geometry, together with their absence in other models, therefore suggests that they are methodological artefacts rather than natural temporal variability. However, the standard deviation alone cannot determine whether the physically plausible variability results from movement of the ocean front, changes in its strength or general ocean warming \citep{garciasuarez2026}.

\subsection{Attention mechanisms}

The attention ablation indicates that factorising attention into separate spatial and temporal operations was beneficial for the imputation task. All three factorised models (space-time, space-time-depth and time-space) achieved the lowest RMSE in both feature mixer conditions, whereas cross- and joint attention generally yielded higher errors and the autoencoder achieved intermediate performance.  
The comparable performance of the three factorised variants further suggests that separating spatial and temporal dimensions was more important than the ordering of the operations or the inclusion of a separate depth-attention step. 

Adding a feature mixer substantially reduced RMSE for all architectures, indicating that its benefit was not specific to a particular attention mechanism. A possible explanation is that the feature mixer enabled information from different features and missingness masks to be combined independently of the restricted spatio-temporal neighbourhoods passed to the attention modules. 

Providing missingness masks to the feature mixer yielded a further, though smaller, reduction in RMSE, indicating that the structure of the observation process contained information relevant to the imputation task. This improvement was not observed for the autoencoder, suggesting that the benefit of missingness information depended on the model architecture. 

Under \SI{100}{\percent} missingness in 1772, cross- and joint attention produced near-uniform, physically unlikely temperature fields since all spatial neighbour tokens were zero-filled and temporal context might have been limited by the graph construction, as the scaling of year, latitude, longitude and depth may have reduced the likelihood of selecting distant but data-rich years such as the 1950s. 
The factorised space-time, space-time-depth and time-space attention models, on the other hand, were able to generate large-scale physical temperature structure even in this sparse regime likely reflecting their ability to exploit temporal information when no spatial context is available. However, no ground truth or simulation data were available in this study for those very early years, preventing any objective evaluation. 

Under partial missingness (2012), sharp discontinuities were most distinct for joint and cross-attention, suggesting a strong focus on point-wise imputation, rather than on generation of coherent physical fields. The factorised models produced considerably smoother structures, possibly because the sequential application of spatial and temporal  attention might have acted as an implicit regulariser: The spatial attention step first aggregates information across a local neighbourhood, generating intermediate representations that were then refined with information from neighbouring years (or vice versa). 
This is consistent with the finding from a video-classification application that divided space-time attention produces more structured representations than joint spatio-temporal attention which was attributed to greater learning capacity~\citep{bertasius2021}.

Due to its best overall performance in RMSE, KGE and CMIP6 MAE, time-space attention with a feature mixer applied to query features and missingness masks was selected as the attention mechanism for SONATI. The remaining visual artefacts under partial missingness were subsequently addressed through masking strategies. 

\subsection{Masking strategies}

Lowest errors (0.030 -- 0.036) were generally achieved by combining one to three masked features per sample with transect or small-to-medium-radius cylindrical masking, suggesting a broad range of well-performing choices. Per-sample masking alone fell within that range, i.e. geographic structure was not strictly necessary to reach this performance band, but it did provide an additional improvement of up to \SI{14.3}{\percent} for the best results. The enhancement through geographic masking did not seem to be an artefact of purely increasing the proportion of effectively hidden values, because the aforementioned best configurations had vastly different fractions (25.2 -- \SI{96.5}{\percent}).
The per-sample and geographical mask combination likely worked well since it reflected missingness patterns in the input data: Often temperature and salinity were measured together, leaving out other features entirely and large geographic areas remained unobserved, informing a block-like masking. Transferring missingness patterns into the masking has been shown to have a positive effect on imputation performance, e.g. by imitating feature-wise missingness proportions~\citep{kim2025}.

Per-feature masking alone resulted in lower RMSE than per-sample masking alone (0.033 vs. 0.035), though the gap was smaller than the seed spread of per-sample masking ($\pm 0.004$), i.e. this difference should be read cautiously. Neither was able to outperform the combined per-sample and geographic masking, supporting the hypothesis that geographically informed masking added a relevant training signal beyond randomness and that per-feature masking might nonetheless be a strong alternative to per-sample masking.

Highest errors were produced by combining per-sample masking with large-radius cylinders ($r \geq 0.7$), by masking four or five features per sample and by geographic masking alone, corresponding to 11.1 -- \SI{100}{\percent} hidden values. One reason is that masking \SI{100}{\percent} removes all observed oceanographic feature context and forces the model to guess. Despite varying fractions of hidden values, the extreme strategies did not seem to be able to reconstruct the values, potentially due to the same explanation, why the moderate combinations worked well: because the resulting masking was not representative enough for the missingness patterns.

RMSE alone did not fully capture imputation quality: The architecture ablation baseline for time-space attention (no feature mixer, mask ratio = 0.3) showed a pronounced background-foreground segregation and transect-like artefacts in the imputed potential temperature field (Figure~E2). Adding the feature mixer (features and missingness information as input) at the same mask ratio removed most of those artefacts and strengthened temporal cohesion, supporting the quantitative RMSE improvement found in the attention ablations through the feature mixer. Combining the feature mixer with the best-performing masking (per-sample and cylindrical) gave a further, smaller improvement in field smoothness, though some patchiness remained. 

Since a larger cluster of masking options resulted in similarly low errors, the best options were given to Optuna for fine-tuning of the masking hyperparameters across all five validation splits. Specifically, Optuna was given the choice between per-feature masking with mask ratios $\in \{0.3, 0.4, 0.5\}$ and per-sample masking (one, two or three features hidden per sample) combined with cylindrical ($r\in\{0.3, 0.4, 0.5\}$) or transect masking ($w\in\{0.02, 0.05\}$). The final best  hyperparameters were identified as masking one feature per sample paired with cylindrical masking with a radius of 0.5. 

\subsection{Uncertainty estimation}

SONATI produced meaningful uncertainty estimates as they were moderately well calibrated and moderately positively correlated with imputation error. This demonstrated that the combination of NLL loss and MC dropout provided a useful signal for identifying difficult predictions. 

The slight underconfidence for all features, i.e. prediction intervals were wider than necessary, was consistent with the z-standard deviations being close to but slightly below one. A potential explanation is the initial min-max scaling, which maps data ranges instead of statistical properties to a common scale across features, potentially distorting variance structure, which the NLL loss learns to predict, a hypothesis requiring verification through comparing against a model trained with standardised features. A possible post-hoc improvement would be recalibration on a held-out dataset~\citep{kuleshov2018}, which rescales the predicted variances to better match empirical coverages. 

Aleatoric uncertainty consistently dominated epistemic uncertainty for all features, with median ratios (pooled across grid cells and seeds) ranging from a factor of about four (salinity) to six (silicate), i.e. SONATI attributed most of its uncertainty to intrinsic noise rather than a lack of knowledge. 
A possible explanation might be a resolution too coarse to capture the natural variability. The applied temporal resolution of 20 years is coarse and cannot represent e.g.  seasonal nutrient cycles. This hypothesis was supported by the ranking of aleatoric dominance, which separated the three nutrients (silicate, phosphate, nitrate with factors of 6.1, 5.6 and 5.5) from the remaining variables (potential temperature, oxygen, salinity with factors of 5.1, 4.8, 4.0). 

Aleatoric uncertainty distributions exhibited pronounced outliers, likely caused by an unconstrained log-variance head. In heteroscedastic NLL training, the network can reduce its loss on difficult points by inflating the predicted variance instead of improving the point estimate~\citep{seitzer2022}. Since aleatoric variance is the exponential of the predicted log-variance, a few such outliers are sufficient to dominate means across grid cells. Therefore, medians across grid cells and seeds were reported for summary statistics, while the uncertainties themselves were averaged over inference passes and seeds. In the current implementation of the installable package \texttt{sonati}, predicted log-variances were clamped to a bounded range. Future work may further consider e.g. a $\beta$-NLL loss~\citep{seitzer2022}. 

Positive uncertainty-missingness correlations indicate that grid cells in sparsely sampled regions received less neighbourhood context and the model correctly reflected this through higher uncertainty, though correlations were weak. 
Oxygen had the strongest correlation between aleatoric uncertainty and missingness ($\rho=0.26 \pm 0.02$), indicating that sparsely sampled regions also tended to be assigned greater irreducible variability. This association may partly reflect unresolved spatial and temporal variability in North Atlantic oxygen, whose distribution is influenced by the oxygen minimum zone in the eastern tropical Atlantic and by ventilation and deep-water formation in the subpolar North Atlantic~\citep{karstensen2008, stendardo2012}. However, this interpretation remains tentative because the observed correlation does not identify the physical cause of the predicted uncertainty.
Phosphate had the highest correlation of epistemic uncertainty with missingness ($\rho=0.28 \pm 0.03$) indicating a lack of model knowledge in data-sparse regions. This leads to the hypothesis that phosphate might be a candidate for further targeted sampling, though additional verification and studies would be required before acting on this finding. 

\section{Conclusion}
This work presented SONATI, a masked  transformer for tabular data imputation, which leverages asymmetric, factorised time-space attention over precomputed neighbourhoods and provides decomposed uncertainties through heteroscedastic NLL loss and multiple inference passes with MC dropout. 
On the gridded North Atlantic dataset and the selected block-structured test set, it achieved the overall best RMSE, KGE and MAE to CMIP6 simulation data and provided moderately calibrated uncertainties. Ablation studies showed that factorised time-space attention and combined geographic and per-sample masking supported accurately and smoothly imputed fields. 

Future work could investigate the performance of SONATI on datasets from other domains and compare it to other models on ungridded data. 
The spatial neighbourhood was constructed in scaled Cartesian coordinates and depth, so that the relative weighting of horizontal and vertical distances was determined by the scaling rather than by physical distance. Constructing the spatial KNN-graph from geographic distances could be explored. 
For the gridded data used here, temporal neighbours were all drawn from the same time step, whereas context from other time steps at the same grid cell entered through the spatial neighbourhood. Future work could restrict temporal neighbourhoods to the same location and spatial neighbourhoods to the same time step.  
A combination of the NLL loss with a reconstruction loss~\citep{du2023} and/or a Fourier imputation loss~\citep{nie2024} might be promising to further improve accuracy and smoothness of predicted fields.
Given the error improvements through tailored masking strategies, another potential direction is the investigation of other masking mechanisms such as copy masking~\citep{gorla2025}. 
Lastly, exploring the value of decomposed uncertainties for downstream tasks such as targeted sampling campaigns would provide application value for oceanographers. 

Data were split under an MAR assumption and further investigation of potential MNAR mechanisms could motivate analyses of imputation robustness to departures from MAR. For example, it could be investigated whether extreme environmental conditions are systematically underrepresented in the observations. Such a bias could arise indirectly because ship-based sampling is concentrated in accessible regions and seasons and often avoids severe weather or sea-ice conditions~\citep{olsen2019}. Very low nutrient concentrations may additionally be affected by analytical detection limits. Comparing observation availability with independent climatological or model-based estimates of environmental extremes could reveal patterns consistent with an MNAR component. However, such comparisons cannot conclusively distinguish MAR from MNAR since the values that are missing are not available for direct verification~\citep{rubin1976}.

\section*{Abbreviations}
\begin{tabular}{ll}
	AWI & Alfred-Wegener-Institut, Helmholtz-Zentrum f\"ur Polar- und Meeresforschung \\
	CMIP6 & Coupled Model Intercomparison Project Phase 6 \\
	COMFORT & Our Common Future Ocean in the Earth System \\
	CSDI & Conditional Score-based Diffusion model for Imputation \\
	DINCAE & Data-Interpolating Convolutional Auto-Encoder \\
	DINEOF & Data INterpolating Empirical Orthogonal Functions \\
	EOF & Empirical orthogonal function \\
	GFDL & Geophysical Fluid Dynamics Laboratory \\
	HPC & High-performance computing \\
	KGE & Kling-Gupta efficiency \\
	KNN & K-nearest neighbours \\
	MAE & Mean absolute error \\
	MAR & Missing at random \\
	MC dropout & Monte Carlo dropout \\
	MCAR & Missing completely at random \\
	ML & Machine learning \\
	MLP & Multi-layer perceptron \\
	MNAR & Missing not at random \\
	NLL & Negative log-likelihood \\
	NOAA & National Oceanic and Atmospheric Administration \\
	ReLU & Rectified linear unit \\
	RMSE & Root mean squared error \\
	SONATI & Spatio-temporal ocean neighbourhood asymmetric transformer for imputation \\
	TPE & Tree-structured Parzen Estimator \\
\end{tabular}

\begin{Backmatter}

\paragraph{Acknowledgments}
We are grateful for the helpful discussions with Claire Monteleoni, Anastase Charantonis and Marvin Wright regarding treatment of time-resolved environmental data sets, missingness mechanisms and imputation strategies. We also thank Gian Giacomo Navarra for the valuable technical assistance with simulation data and paper structure. 
This work has been kindly supported by computing infrastructure of 
the Alfred-Wegener Institut, Helmholtz-Zentrum f\"ur Polar- und Meeresforschung. 

\paragraph{Funding Statement}
YJ has been funded through the University of Bremen, the Alfred-Wegener-Institut, Helmholtz-Zentrum für Polar- und Meeresforschung and 
the Helmholtz School for Marine Data Science (MarDATA).
BPK, MS and SM were supported by their employing institutions. The funders had no role in the study design, data collection and analysis, decision to publish or preparation of the manuscript.

\paragraph{Competing Interests}
None.

\paragraph{Data Availability Statement}
\label{sec:data_availability}
\textit{Input data.} The observational data are from the COMFORT dataset~\citep{comfort_dataset2022}. Its pre-processing can be reproduced with the Python library \texttt{comfort-db} (\cite{jenniges2026}; version 0.1.0, example 9). The CMIP6 GFDL-ESM4 historical simulation~\citep{cmip2018} was accessed through the Pangeo CMIP6 cloud catalogue (\url{https://pangeo-data.github.io/pangeo-cmip6-cloud/}).

\textit{Code.} The experiment code is hosted on GitHub and archived on Zenodo~\citep{jenniges2026_experiment_code}. The final SONATI model is available as the Python package \texttt{sonati} on PyPI~\citep{jenniges2026_sonati_code}.

\textit{Output data.} The imputed North Atlantic dataset is openly available on Zenodo~\citep{jenniges2026_sonati_data}.

\paragraph{Ethical Standards}
The research meets all ethical guidelines, including adherence to the legal requirements of the study country.

\paragraph{Author Contributions}
Conceptualisation: YJ, BPK, MS, SM.
Methodology: YJ, BPK, SM.
Software: YJ.
Formal analysis: YJ, BPK.
Investigation: YJ.
Visualisation: YJ, BPK.
Writing -- original draft: YJ.
Writing -- review and editing: YJ, BPK, MS, SM.
Funding acquisition: YJ, BPK, SM.
Project administration: YJ.
Supervision: BPK, SM.

\paragraph{Generative AI usage declaration}
The authors used generative artificial intelligence (GAI) tools (ChatGPT, OpenAI, version not recorded; Claude Code, Anthropic, Claude Sonnet 5 and 5.5 models) to improve the clarity of expressions in the manuscript, detect and correct code errors, to improve code structure, plotting functions and documentation and to discuss experimental design. Moreover, GAI was applied to wrap the final SONATI model into a standalone, installable PyPI package and add tests (\texttt{sonati}). All AI-assisted outputs were carefully reviewed, verified and edited line-by-line by the authors, who take full responsibility for the content and accuracy of this work.


\printbibliography

\end{Backmatter}

\clearpage
\section*{Supplementary material}
\raggedbottom

\begin{appendix}
\counterwithin{figure}{section}
\renewcommand{\thefigure}{\thesection\arabic{figure}}
\counterwithin{table}{section}
\renewcommand{\thetable}{\thesection\arabic{table}}
\section{Time resolution}
\label{appendixA}

\begin{figure}[!htbp]
	\begin{center}
		\includegraphics[width=0.85\textwidth]{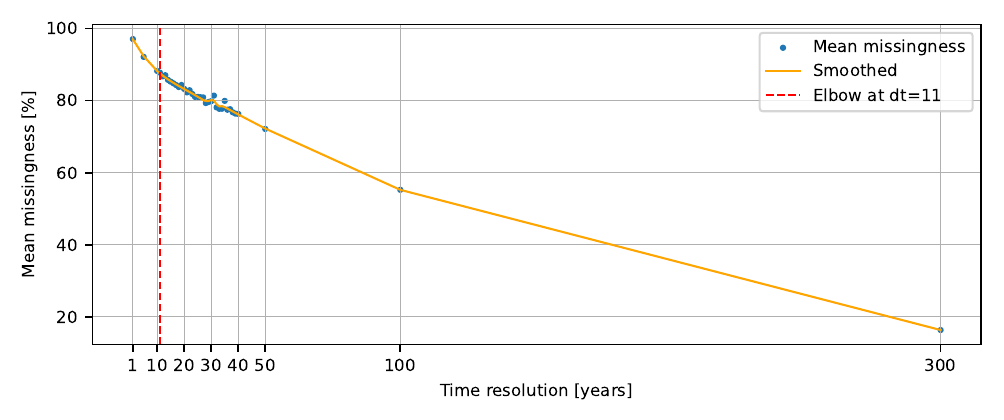}
		\caption{Mean missingness, i.e. the percentage of the six parameters missing in a grid cell averaged over grid cells, time steps and  parameters, versus time-bin width as blue points. The orange line displays a fitted, smooth spline (UnivariateSpline, Python library scipy, version 1.16.2, smoothing factor $s=10$) on which the elbow is computed (Python library kneed, version 0.8.5, concave and decreasing curve, default sensitivity $S=1$). The elbow at an 11-year time step size represents the trade-off between data availability and time resolution}
		\label{app:fig:time_resolution_vs_missingness}
	\end{center}
\end{figure}

\begin{figure}[!htbp]
	\begin{center}
		\includegraphics[width=0.85\textwidth]{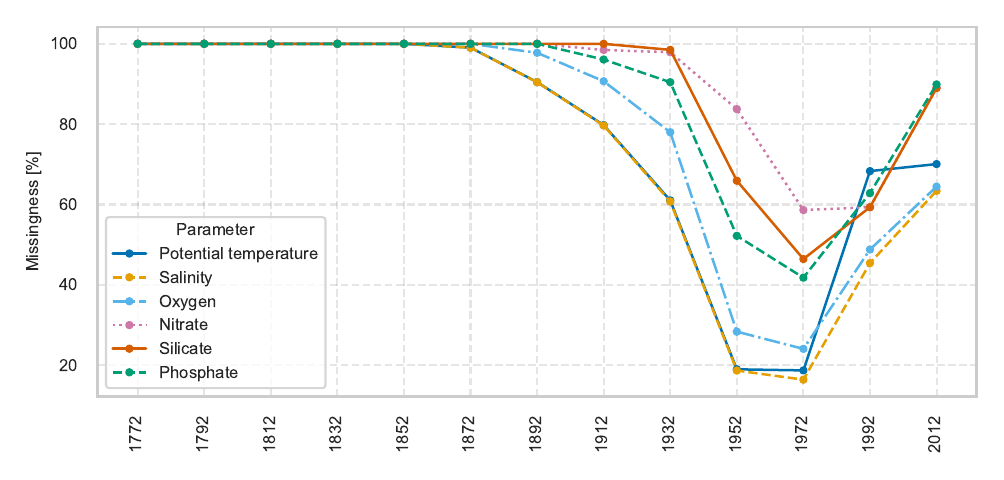}
		\caption{Per-parameter missingness per 20-year time step, i.e. the percentage of grid cells with no measurement of the parameter. Salinity was generally best sampled, closely followed by temperature, while nutrient data was sparser. Missingness was lowest for all features in the time step starting in the year 1972}
		\label{app:fig:missingness_over_time}
	\end{center}
\end{figure}

\section{Data analysis}
\label{app:sec:data_analysis}
\begin{figure}[H]
	\begin{center}
		\includegraphics[width=\textwidth]{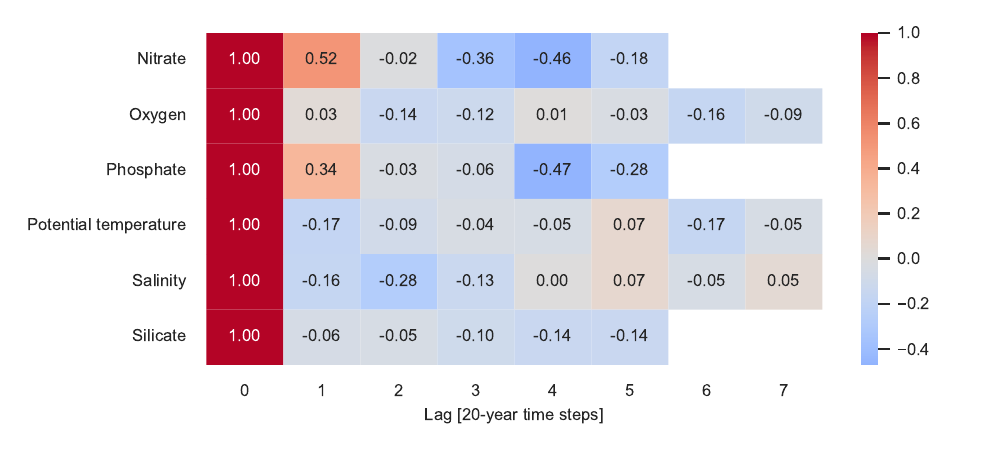}
		\caption{Temporal autocorrelations for six oceanographic parameters in the North Atlantic. 
		Each cell shows the autocorrelation at a 20-year lag. 
		For the nutrients, lags without a value occurred as fewer time steps contained respective data.  
		Temperature, salinity and oxygen decorrelate rapidly indicating limited temporal persistence at the considered resolution. 
		Nutrients show slightly stronger short-lag correlations and oscillations at medium lags, suggesting modest multi-decadal correlation}
		\label{app:fig:temporal_autocorrelations}
	\end{center}
\end{figure}

When splitting the data using block holdout, an explicit choice of block size, shape and assignment is required. 
Previous work indicates that block size is the dominant factor and should be guided, among other factors, by the application and spatial and/or temporal autocorrelation ranges, i.e. the distances over which dependence decays~\citep{stock2025}. 
Here, horizontal decorrelation varied between approximately \SI{6200}{\kilo\metre} and \SI{10900}{\kilo\metre} corresponding to half to full basin scales (Table~\ref{app:tab:variogram}). 
Using such large spatial blocks would leave very little data for training in the given sparse data setting. 
Therefore, smaller horizontal block sizes were investigated. Temporal decorrelation ranged between 3 and 140 years, where values below the 20-year grid resolution indicate decorrelation within the first lag, with non-monotonic trends, possibly due to strong temporal sparsity.
An additional correlogram analysis was therefore performed (Figure~\ref{app:fig:temporal_autocorrelations}), revealing low correlations after one lag for temperature, salinity and oxygen and medium ones for the nutrients nearly vanishing after one more lag. 
Due to temporal sparsity and correlogram results, the block size was fixed to 20 years, corresponding to the native temporal grid resolution.

\begin{table}[H]
	\centering
	\caption{Summary of variogram analysis (Python library scikit-gstat, version 1.0.21) of six oceanographic parameters in the North Atlantic. 
	Variograms were computed on 10,000 randomly drawn observed grid cells per parameter from the 20-year gridded dataset using unscaled values and separately for horizontal space (latitude and longitude), depth and time. 
	Horizontal coordinates were transformed to kilometres using a Lambert azimuthal equal-area projection, while depth is reported in metres and temporal lags in years. 
	The variogram sill (plateau semivariance) and nugget (unresolved small-scale variability and measurement noise) are given in squared parameter units ([\SI{}{\celsius}]$^2$, [psu]$^2$, [\SI{}{\micro\mole\per\kilo\gram}]$^2$). The range is the lag distance at which the semivariance approaches the sill}
	\label{app:tab:variogram}
	\begin{tabular}{l|rrrrrr}
		\toprule
		\multicolumn{1}{l}{} & Potential temperature & Salinity & Oxygen & Nitrate & Silicate & Phosphate \\
		\midrule
		Spatial range [km] & 6246 & 10869 & 7845 & 7051 & 10631 & 10860 \\
		Spatial sill & 42.0 & 23.9 & 4956.1 & 61.2 & 297.7 & 0.7 \\
		Spatial nugget & 11.2 & 0.0 & 0.0 & 58.2 & 25.8 & 0.0 \\
		Depth range [m] & 2192 & 358 & 4000 & 303 & 4000 & 4000 \\
		Depth sill & 20.8 & 0.0 & 0.0 & 0.1 & 435.6 & 0.0 \\
		Depth nugget & 31.7 & 6.2 & 2892.5 & 93.8 & 0.0 & 0.4 \\
		Time range [years] & 140 & 140 & 140 & 100 & 80 & 3 \\
		Time sill & 0.0 & 17.2 & 0.0 & 0.0 & 54.1 & 0.0 \\
		Time nugget & 41.5 & 2.1 & 3294.9 & 102.3 & 144.9 & 0.4 \\
		\bottomrule
	\end{tabular}
\end{table}

\begin{figure}[H]
	\begin{center}
		\includegraphics[width=\textwidth]{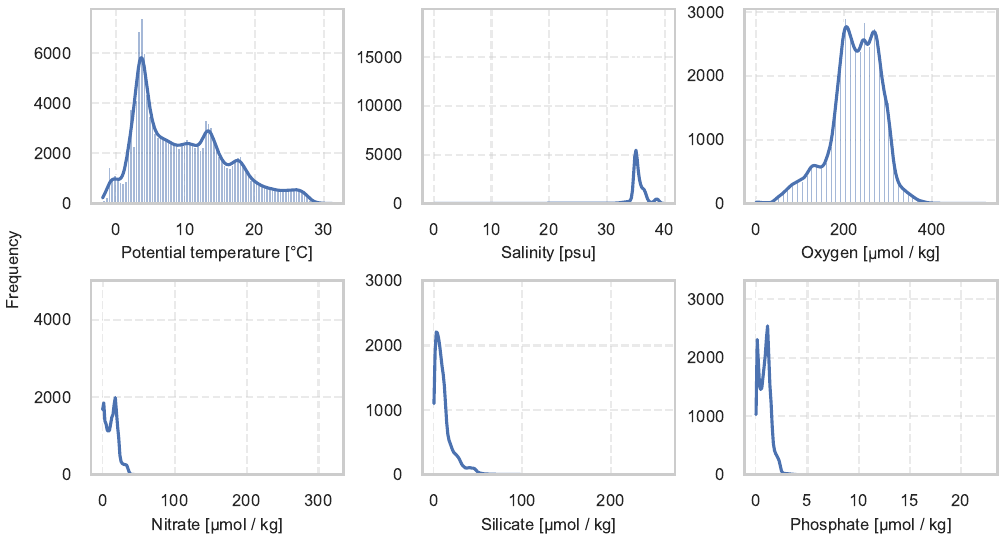}
		\caption{Histograms of all observed values per parameter with kernel density estimates (bin numbers set automatically by \texttt{seaborn}, version 0.13.2) in the 20-year gridded dataset (637,390 grid cells, of which 55,349 to 159,907 have a value per parameter). The long upper tails of nitrate, silicate and phosphate mostly originate from a few extreme values in river-influenced coastal waters and the Black Sea}
		\label{app:fig:histograms}
	\end{center}
\end{figure}

\section{Model tuning}
\begin{longtable}{p{2.8cm} p{3.5cm} >{\raggedright\arraybackslash}p{6.5cm}}
	\caption{Hyperparameters and their ranges explored in the Optuna tuning studies. Comma-separated lists indicate discrete values, while dash-separated numbers indicate a continuous range that Optuna optimisation could choose from} 
	\label{app:tab:tuning}\\
	\toprule
	\textbf{Model} & \textbf{Hyperparameter} & \textbf{Values} \\
	\midrule
	\endfirsthead
	\toprule
	\textbf{Model} & \textbf{Hyperparameter} & \textbf{Values} \\
	\midrule
	\endhead
	\midrule
	\multicolumn{3}{r}{\textit{Continued on next page}} \\
	\endfoot
	\bottomrule
	\endlastfoot
	
	SimpleImputer & strategy & mean, median \\
	\midrule
	
	\multirow{2}{*}{KNN}
	& n\_neighbors & 1 -- 50 \\
	& weights      & uniform, distance \\
	\midrule
	
	\multirow{2}{*}{MissForest}
	& n\_estimators & 50 -- 200 \\
	& max\_features & None, sqrt, 0.5 \\
	\midrule
	
	\multirow{8}{*}{ReMasker}
	& batch\_size     & 64, 128, 256, 512 \\
	& mask\_ratio     & 0.0 -- 0.99 \\
	& embed\_dim      & 16, 32, 64, 128, 256, 512 \\
	& depth           & 2 -- 8 \\
	& decoder\_depth  & 1 -- 8 \\
	& num\_heads      & 2, 4, 8 \\
	& encode\_func    & linear, active \\
	& max\_epochs     & 20 \\
	\midrule
	
	\multirow{8}{*}{MLP}
	& loss          & hetero \\
	& patience      & 10 \\
	& n\_epochs     & 80 \\
	& batch\_size   & 128, 256, 512 \\
	& learning\_rate & $10^{-5}$ -- $10^{-2}$ (log scale) \\
	& mask\_ratio   & 0.0 -- 0.99 \\
	& dropout       & 0.1 -- 0.4 \\
	& hidden\_dims  & [64,~64], [128,~64], [256,~128], [128,~64,~32],
	[256,~128,~64], [256,~128,~64,~32], [512,~256,~128],
	[512,~256,~128,~64], [256,~256,~128,~64],
	[128,~128,~64,~32,~16] \\
	\midrule
	
	\multirow{15}{*}{SONATI}
	& loss                  & hetero \\
	& patience              & 5 \\
	& n\_epochs             & 20 \\
	& n\_neighbours         & 5 -- 60 \\
	& batch\_size           & 64, 128, 256, 512 \\
	& learning\_rate        & $10^{-5}$ -- $10^{-3}$ (log scale) \\
	& mask\_ratio           & 0.3, 0.4, 0.5 \\
	& masking\_strategies   & [per-sample], [per-sample, cylinder], [per-sample, transect], [per-feature] \\
	& cylinder\_mask\_radius  & 0.3, 0.4, 0.5 \\
	& transect\_mask\_width & 0.02, 0.05 \\
	& d\_model              & 32, 64, 128, 256, 512 \\
	& nhead                 & 2, 4, 8 \\
	& nlayers               & 2 -- 8 \\
	& dim\_feedforward      & 128, 256, 512, 1024 \\
	& dropout               & 0.1 -- 0.4 \\
\end{longtable}

\begingroup\setlength{\tabcolsep}{2pt}
\begin{longtable}{p{2.0cm} p{3.0cm} p{2.9cm} p{1.5cm} p{1.6cm} p{2.4cm}}
	\caption{Best hyperparameter combinations for each imputation model obtained from Optuna studies. The hyperparameters are sorted according to their hyperparameter importance, which was computed with Optuna's fANOVA implementation. No importance (--) was reported for cylinder\_mask\_radius due to the way the study was set up. 
	The validation RMSE was computed in $[0,1]$-scaled space.
	The duration measures the time per trial, which consisted of training the model on each of the five train-validation splits. Note that the standard models were tuned on CPU nodes of a  high-performance system, while the neural networks were tuned on GPU nodes}
	\label{app:tab:hyps}\\
	\toprule
	\textbf{Model} & \textbf{Hyperparameters} & \textbf{Values} & \textbf{Validation RMSE} & \textbf{Duration [hh:mm:ss]} & \textbf{Hyperparameter importance} \\
	\midrule
	\endfirsthead
	\toprule
	\textbf{Model} & \textbf{Hyperparameters} & \textbf{Values} & \textbf{Validation RMSE} & \textbf{Duration [hh:mm:ss]} & \textbf{Hyperparameter importance} \\
	\midrule
	\endhead
	\midrule
	\multicolumn{6}{r}{\textit{Continued on next page}} \\
	\endfoot
	\bottomrule
	\endlastfoot
	
	SimpleImputer
	& strategy
	& mean
	& 0.12094
	& 00:00:04
	& 1 \\
	\midrule
	
	KNN
	& \makecell[l]{n\_neighbors \\ weights}
	& \makecell[l]{5 \\ distance}
	& 0.03716
	& 20:27:39
	& \makecell[l]{0.899 \\ 0.101} \\
	\midrule
	
	MissForest
	& \makecell[l]{max\_features \\ n\_estimators}
	& \makecell[l]{sqrt \\ 178}
	& 0.02826
	& 00:06:02
	& \makecell[l]{0.994 \\ 0.006} \\
	\midrule
	
	\makecell[l]{ReMasker}
	& \makecell[l]{mask\_ratio \\ embed\_dim \\ depth \\ encode\_func \\ decoder\_depth \\ batch\_size \\ num\_heads}
	& \makecell[l]{0.5168 \\ 64 \\ 2 \\ linear \\ 5 \\ 128 \\ 4}
	& 0.04007
	& 05:16:02
	& \makecell[l]{0.696 \\ 0.241 \\ 0.033 \\ 0.011 \\ 0.011 \\ 0.007 \\ 0.001} \\
	\midrule
	
	MLP
	& \makecell[l]{learning\_rate \\ dropout \\ hidden\_dims \\ mask\_ratio \\ batch\_size}
	& \makecell[l]{$1.45 \times 10^{-4}$ \\ 0.1137 \\ {[512, 256, 128]} \\ 0.68 \\ 128}
	& 0.03022
	& 01:33:28
	& \makecell[l]{0.506 \\ 0.383 \\ 0.053 \\ 0.041 \\ 0.017} \\
	\midrule
	
	SONATI
	& \makecell[l]{learning\_rate \\ batch\_size \\ d\_model \\ dropout \\ dim\_feedforward \\ n\_neighbours \\ nhead \\ nlayers \\ mask\_ratio \\ masking\_strategies \\ cylinder\_mask\_radius}
	& \makecell[l]{$6.851 \times 10^{-5}$ \\ 64 \\ 512 \\ 0.2089 \\ 512 \\ 36 \\ 8 \\ 5 \\ 0.3 \\ {[per-sample, cylinder]} \\ 0.5}
	& 0.02583
	& 15:01:23
	& \makecell[l]{0.214 \\ 0.197 \\ 0.151 \\ 0.110 \\ 0.095 \\ 0.055 \\ 0.053 \\ 0.048 \\ 0.040 \\ 0.037 \\ --} \\
\end{longtable}
\endgroup

\section{Model testing}
\begin{table}[H]
	\centering
	\caption{Kling-Gupta efficiency (KGE) per $[0,1]$-scaled parameter for each model (mean $\pm$ standard deviation across five seeds). KNN and mean imputation are deterministic and therefore have no spread}
	\label{app:tab:kge_per_var}
	\resizebox{\textwidth}{!}{%
		\begin{tabular}{lllllll}
			\toprule
			Model & Potential temperature & Salinity & Oxygen & Nitrate & Silicate & Phosphate \\
			\midrule
			MLP & 0.40 $\pm$ 0.32 & 0.50 $\pm$ 0.11 & 0.52 $\pm$ 0.12 & $-$6.27 $\pm$ 7.69 & $-$53.89 $\pm$ 120.55 & $-$48.62 $\pm$ 108.46 \\
			Mean & $-$0.42 & $-$0.41 & $-$0.41 & $-$0.41 & $-$0.41 & $-$0.41 \\
			KNN & 0.92 & 0.87 & 0.89 & 0.84 & 0.65 & 0.76 \\
			MissForest & 0.97 $\pm$ 0.00 & 0.97 $\pm$ 0.00 & \textbf{0.96 $\pm$ 0.00} & 0.83 $\pm$ 0.01 & \textbf{0.72 $\pm$ 0.02} & 0.75 $\pm$ 0.02 \\
			ReMasker & 0.97 $\pm$ 0.00 & \textbf{0.98 $\pm$ 0.01} & 0.92 $\pm$ 0.01 & 0.75 $\pm$ 0.02 & 0.69 $\pm$ 0.02 & 0.69 $\pm$ 0.03 \\
			SONATI & \textbf{0.98 $\pm$ 0.00} & 0.97 $\pm$ 0.02 & 0.95 $\pm$ 0.01 & \textbf{0.91 $\pm$ 0.02} & 0.69 $\pm$ 0.02 & \textbf{0.82 $\pm$ 0.04} \\
			\bottomrule
		\end{tabular}
	}
\end{table}

\begin{table}[H]
	\centering
	\caption{Mean absolute differences of the imputed fields with CMIP6 simulation data for each model in physical units across five seeds. KNN and mean imputation are deterministic and therefore have no spread}
	\label{app:tab:cmip_mae_per_var_unscaled}
	\resizebox{\textwidth}{!}{%
		\begin{tabular}{lllllll}
			\toprule
			Model & Potential temperature [\SI{}{\celsius}] & Salinity [psu] & Oxygen [\SI{}{\micro\mole\per\kilo\gram}] & Nitrate [\SI{}{\micro\mole\per\kilo\gram}] & Silicate [\SI{}{\micro\mole\per\kilo\gram}] & Phosphate [\SI{}{\micro\mole\per\kilo\gram}] \\
			\midrule
			MLP & 1.298 $\pm$ 0.103 & 0.269 $\pm$ 0.011 & \textbf{18.925 $\pm$ 0.217} & 3.714 $\pm$ 0.311 & 6.629 $\pm$ 0.213 & 0.233 $\pm$ 0.007 \\
			Mean & 5.558 & 0.646 & 41.650 & 5.993 & 11.779 & 0.408 \\
			KNN & 1.634 & 0.354 & 24.195 & 4.891 & 9.923 & 0.289 \\
			MissForest & \textbf{1.107 $\pm$ 0.013} & 0.280 $\pm$ 0.005 & 20.907 $\pm$ 0.166 & 3.507 $\pm$ 0.069 & 9.573 $\pm$ 0.496 & 0.237 $\pm$ 0.002 \\
			ReMasker & 1.209 $\pm$ 0.071 & 0.305 $\pm$ 0.017 & 20.050 $\pm$ 0.556 & \textbf{3.442 $\pm$ 0.495} & 7.510 $\pm$ 0.272 & \textbf{0.227 $\pm$ 0.011} \\
			SONATI & 1.135 $\pm$ 0.014 & \textbf{0.262 $\pm$ 0.010} & 20.203 $\pm$ 0.275 & 3.529 $\pm$ 0.033 & \textbf{6.200 $\pm$ 0.141} & 0.231 $\pm$ 0.013 \\
			\bottomrule
		\end{tabular}
	}
\end{table}

\begin{table}[H]
	\centering
	\caption{Mean absolute differences of the imputed $[0,1]$-scaled fields with CMIP6 simulation data for each model across five seeds. KNN and mean imputation are deterministic and therefore have no spread}
	\label{app:tab:cmip_mae_per_var_scaled}
	\resizebox{\textwidth}{!}{%
		\begin{tabular}{lllllll}
			\toprule
			Model & Potential temperature & Salinity & Oxygen & Nitrate & Silicate & Phosphate \\
			\midrule
			MLP & 0.0392 $\pm$ 0.0031 & 0.0067 $\pm$ 0.0003 & \textbf{0.0431 $\pm$ 0.0005} & 0.0116 $\pm$ 0.0010 & 0.0255 $\pm$ 0.0008 & 0.0104 $\pm$ 0.0003 \\
			Mean & 0.1679 & 0.0162 & 0.0948 & 0.0188 & 0.0453 & 0.0181 \\
			KNN & 0.0494 & 0.0089 & 0.0551 & 0.0153 & 0.0381 & 0.0129 \\
			MissForest & \textbf{0.0335 $\pm$ 0.0004} & 0.0070 $\pm$ 0.0001 & 0.0476 $\pm$ 0.0004 & 0.0110 $\pm$ 0.0002 & 0.0368 $\pm$ 0.0019 & 0.0106 $\pm$ 0.0001 \\
			ReMasker & 0.0365 $\pm$ 0.0022 & 0.0077 $\pm$ 0.0004 & 0.0457 $\pm$ 0.0013 & \textbf{0.0108 $\pm$ 0.0016} & 0.0289 $\pm$ 0.0010 & \textbf{0.0101 $\pm$ 0.0005} \\
			SONATI & 0.0343 $\pm$ 0.0004 & \textbf{0.0066 $\pm$ 0.0002} & 0.0460 $\pm$ 0.0006 & 0.0111 $\pm$ 0.0001 & \textbf{0.0238 $\pm$ 0.0005} & 0.0103 $\pm$ 0.0006 \\
			\bottomrule
		\end{tabular}
	}
\end{table}

\section{Ablations}
\begin{sidewaystable}
	\centering
	\caption{Performance of models aggregated across five seeds with various attention types and feature mixer setups, evaluated overall and, for root mean squared error (RMSE), per parameter in $[0,1]$-normalised space. 
	Overall RMSE was computed across all test values and hence weights parameters by their number of observations. Overall Kling-Gupta efficiency (KGE) was calculated as the median across the per-parameter KGEs and the overall mean absolute error (MAE) to CMIP6 simulation data as average across per-parameter MAEs}
	\label{app:tab:attention}
	\resizebox{\textheight}{!}{%
		\begin{tabular}{lllllllllll}
			\toprule
			\multicolumn{2}{l}{Model} & RMSE $\downarrow$ & KGE $\uparrow$ & CMIP MAE $\downarrow$ & \multicolumn{6}{c}{RMSE per parameter $\downarrow$} \\
			\cmidrule(lr){6-11}
			Attention type & Feature mixer &  &  &  & Potential temperature & Salinity & Oxygen & Nitrate & Silicate & Phosphate \\
			\midrule
			Autoencoder & Features & 0.0602 $\pm$ 0.0079 & 0.73 $\pm$ 0.06 & 0.0301 $\pm$ 0.0042 & 0.0879 $\pm$ 0.0116 & 0.0499 $\pm$ 0.0127 & 0.0642 $\pm$ 0.0051 & 0.0119 $\pm$ 0.0021 & 0.0378 $\pm$ 0.0018 & 0.0137 $\pm$ 0.0013 \\
			Autoencoder & Features+masks & 0.0608 $\pm$ 0.0038 & 0.74 $\pm$ 0.02 & 0.0278 $\pm$ 0.0022 & 0.0844 $\pm$ 0.0149 & 0.0577 $\pm$ 0.0074 & 0.0628 $\pm$ 0.0041 & 0.0123 $\pm$ 0.0015 & 0.0378 $\pm$ 0.0019 & 0.0146 $\pm$ 0.0018 \\
			Autoencoder & None & 0.0778 $\pm$ 0.0086 & 0.66 $\pm$ 0.06 & 0.0462 $\pm$ 0.0057 & 0.1123 $\pm$ 0.0077 & 0.0684 $\pm$ 0.0212 & 0.0820 $\pm$ 0.0156 & 0.0133 $\pm$ 0.0006 & 0.0376 $\pm$ 0.0030 & 0.0157 $\pm$ 0.0023 \\
			\midrule
			Cross-attention & Features & 0.0921 $\pm$ 0.0188 & 0.66 $\pm$ 0.05 & 0.0457 $\pm$ 0.0085 & 0.1402 $\pm$ 0.0416 & 0.0700 $\pm$ 0.0146 & 0.0962 $\pm$ 0.0153 & 0.0131 $\pm$ 0.0015 & 0.0420 $\pm$ 0.0023 & 0.0165 $\pm$ 0.0005 \\
			Cross-attention & Features+masks & 0.0852 $\pm$ 0.0154 & 0.67 $\pm$ 0.12 & 0.0340 $\pm$ 0.0036 & 0.1428 $\pm$ 0.0325 & 0.0572 $\pm$ 0.0179 & 0.0697 $\pm$ 0.0077 & 0.0130 $\pm$ 0.0020 & 0.0374 $\pm$ 0.0034 & 0.0153 $\pm$ 0.0036 \\
			Cross-attention & None & 0.1406 $\pm$ 0.0379 & 0.50 $\pm$ 0.08 & 0.0830 $\pm$ 0.0126 & 0.1714 $\pm$ 0.0310 & 0.1718 $\pm$ 0.0797 & 0.1348 $\pm$ 0.0269 & 0.0150 $\pm$ 0.0015 & 0.0440 $\pm$ 0.0070 & 0.0172 $\pm$ 0.0016 \\
			\midrule
			Joint attention & Features & 0.0814 $\pm$ 0.0135 & 0.72 $\pm$ 0.05 & 0.0418 $\pm$ 0.0031 & 0.1398 $\pm$ 0.0301 & 0.0367 $\pm$ 0.0097 & 0.0740 $\pm$ 0.0104 & 0.0109 $\pm$ 0.0015 & 0.0380 $\pm$ 0.0022 & 0.0138 $\pm$ 0.0014 \\
			Joint attention & Features+masks & 0.0543 $\pm$ 0.0051 & 0.73 $\pm$ 0.02 & 0.0357 $\pm$ 0.0067 & 0.0829 $\pm$ 0.0112 & 0.0413 $\pm$ 0.0064 & 0.0543 $\pm$ 0.0080 & 0.0116 $\pm$ 0.0015 & \textbf{0.0350 $\pm$ 0.0025} & 0.0137 $\pm$ 0.0020 \\
			Joint attention & None & 0.1028 $\pm$ 0.0089 & 0.73 $\pm$ 0.06 & 0.0594 $\pm$ 0.0065 & 0.1740 $\pm$ 0.0173 & 0.0669 $\pm$ 0.0234 & 0.0846 $\pm$ 0.0137 & 0.0108 $\pm$ 0.0016 & 0.0364 $\pm$ 0.0010 & 0.0125 $\pm$ 0.0007 \\
			\midrule
			Space-time attention & Features & 0.0524 $\pm$ 0.0077 & 0.81 $\pm$ 0.03 & 0.0278 $\pm$ 0.0041 & 0.0878 $\pm$ 0.0189 & \textbf{0.0207 $\pm$ 0.0024} & 0.0499 $\pm$ 0.0038 & 0.0091 $\pm$ 0.0012 & 0.0376 $\pm$ 0.0009 & 0.0122 $\pm$ 0.0011 \\
			Space-time attention & Features+masks & 0.0489 $\pm$ 0.0095 & 0.81 $\pm$ 0.03 & 0.0252 $\pm$ 0.0028 & 0.0789 $\pm$ 0.0214 & 0.0277 $\pm$ 0.0084 & 0.0463 $\pm$ 0.0050 & 0.0093 $\pm$ 0.0008 & 0.0353 $\pm$ 0.0016 & 0.0121 $\pm$ 0.0009 \\
			Space-time attention & None & 0.0702 $\pm$ 0.0111 & 0.76 $\pm$ 0.07 & 0.0340 $\pm$ 0.0032 & 0.1172 $\pm$ 0.0226 & 0.0450 $\pm$ 0.0115 & 0.0609 $\pm$ 0.0087 & 0.0098 $\pm$ 0.0019 & 0.0357 $\pm$ 0.0026 & 0.0124 $\pm$ 0.0013 \\
			\midrule
			Space-time-depth attention & Features & 0.0499 $\pm$ 0.0046 & 0.79 $\pm$ 0.02 & 0.0258 $\pm$ 0.0020 & 0.0769 $\pm$ 0.0126 & 0.0295 $\pm$ 0.0040 & 0.0537 $\pm$ 0.0037 & 0.0088 $\pm$ 0.0009 & 0.0388 $\pm$ 0.0020 & 0.0126 $\pm$ 0.0014 \\
			Space-time-depth attention & Features+masks & 0.0449 $\pm$ 0.0055 & 0.81 $\pm$ 0.03 & 0.0254 $\pm$ 0.0019 & \textbf{0.0604 $\pm$ 0.0148} & 0.0378 $\pm$ 0.0144 & 0.0499 $\pm$ 0.0048 & 0.0091 $\pm$ 0.0015 & 0.0357 $\pm$ 0.0023 & 0.0120 $\pm$ 0.0010 \\
			Space-time-depth attention & None & 0.0631 $\pm$ 0.0059 & 0.78 $\pm$ 0.03 & 0.0320 $\pm$ 0.0026 & 0.1065 $\pm$ 0.0140 & 0.0258 $\pm$ 0.0049 & 0.0621 $\pm$ 0.0008 & \textbf{0.0079 $\pm$ 0.0005} & 0.0373 $\pm$ 0.0009 & \textbf{0.0109 $\pm$ 0.0004} \\
			\midrule
			Time-space attention & Features & 0.0447 $\pm$ 0.0049 & 0.82 $\pm$ 0.02 & 0.0247 $\pm$ 0.0016 & 0.0648 $\pm$ 0.0114 & 0.0289 $\pm$ 0.0086 & 0.0516 $\pm$ 0.0034 & 0.0092 $\pm$ 0.0011 & 0.0369 $\pm$ 0.0013 & 0.0128 $\pm$ 0.0014 \\
			Time-space attention & Features+masks & \textbf{0.0419 $\pm$ 0.0057} & \textbf{0.84 $\pm$ 0.04} & \textbf{0.0232 $\pm$ 0.0017} & 0.0622 $\pm$ 0.0129 & 0.0274 $\pm$ 0.0083 & \textbf{0.0451 $\pm$ 0.0046} & 0.0083 $\pm$ 0.0009 & 0.0357 $\pm$ 0.0012 & 0.0117 $\pm$ 0.0009 \\
			Time-space attention & None & 0.0646 $\pm$ 0.0156 & 0.79 $\pm$ 0.04 & 0.0355 $\pm$ 0.0064 & 0.0892 $\pm$ 0.0146 & 0.0629 $\pm$ 0.0447 & 0.0579 $\pm$ 0.0018 & 0.0086 $\pm$ 0.0006 & 0.0368 $\pm$ 0.0016 & 0.0119 $\pm$ 0.0006 \\
			\bottomrule
		\end{tabular}
	}
\end{sidewaystable}

\begin{figure}[H]
	\centering
	\includegraphics[width=\textwidth]{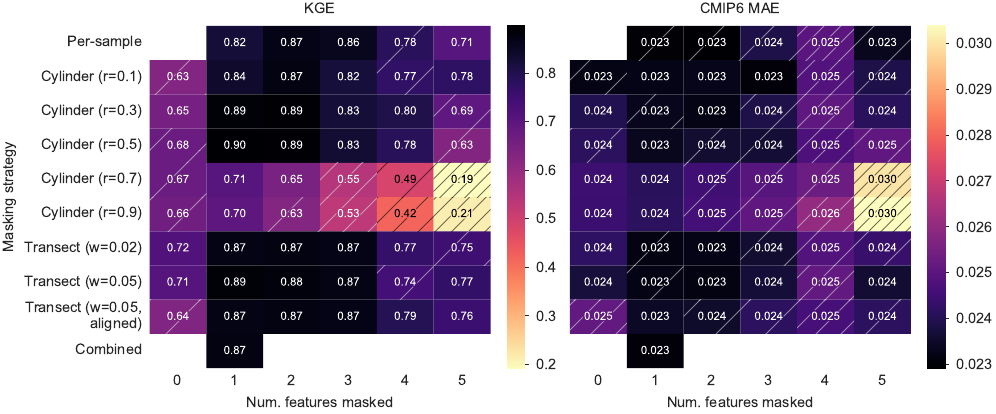}
	\caption{Kling-Gupta efficiency (KGE, median across parameters, left) and mean absolute error (MAE) to CMIP6 simulation data (right) of time-space attention under various masking strategies, applied alone and in combination, evaluated on the \SI{15}{\degree}-block validation set. Columns denote the number of features masked per data point in addition to the row's masking strategy. The first column corresponds to that strategy applied on its own, without extra per-sample masking. The first row shows per-sample masking alone. The middle rows combine per-sample masking with cylindrical masking  (spatial radius $r\in\{0.1, 0.3, 0.5, 0.7, 0.9\}$), simulating regional data gaps and transect masking (width $w\in\{0.02, 0.05\}$), simulating ship-track gaps, applied in a random or grid-aligned direction. The last row shows per-sample, transect and cylindrical masking simultaneously. Cell colour encodes the magnitude of the respective error metric, while hatching indicates the coefficient of variation across the five seeds for that cell, min-max normalised within the heatmap. Denser hatching marks a less stable metric across seeds}
	\label{app:fig:masking_strategies}
\end{figure}

\begin{table}[H]
	\centering
	\caption{Performance of different mask ratios for per-feature masking on a time-space factorised transformer across five seeds. KGE is the median across parameters}
	\label{app:tab:per_feature_masking}
	\begin{tabular}{llll}
		\toprule
		Mask ratio & RMSE $\downarrow$ & KGE $\uparrow$ & CMIP MAE $\downarrow$ \\
		\midrule
		0.1 & 0.0400 $\pm$ 0.0028 & 0.82 $\pm$ 0.04 & \textbf{0.0228 $\pm$ 0.0010} \\
		0.2 & 0.0338 $\pm$ 0.0019 & \textbf{0.88 $\pm$ 0.03} & 0.0230 $\pm$ 0.0004 \\
		0.3 & 0.0333 $\pm$ 0.0018 & \textbf{0.88 $\pm$ 0.02} & \textbf{0.0228 $\pm$ 0.0009} \\
		0.4 & 0.0330 $\pm$ 0.0020 & 0.87 $\pm$ 0.02 & 0.0236 $\pm$ 0.0005 \\
		0.5 & 0.0330 $\pm$ 0.0019 & 0.86 $\pm$ 0.02 & 0.0232 $\pm$ 0.0003 \\
		0.6 & \textbf{0.0326 $\pm$ 0.0019} & \textbf{0.88 $\pm$ 0.02} & 0.0233 $\pm$ 0.0007 \\
		0.7 & 0.0345 $\pm$ 0.0036 & 0.86 $\pm$ 0.04 & 0.0239 $\pm$ 0.0007 \\
		0.8 & 0.0361 $\pm$ 0.0030 & 0.82 $\pm$ 0.07 & 0.0236 $\pm$ 0.0006 \\
		0.9 & 0.0383 $\pm$ 0.0058 & 0.79 $\pm$ 0.10 & 0.0234 $\pm$ 0.0003 \\
		\bottomrule
	\end{tabular}
\end{table}

\begin{figure}[H]
	\centering
	\includegraphics[width=\textwidth]{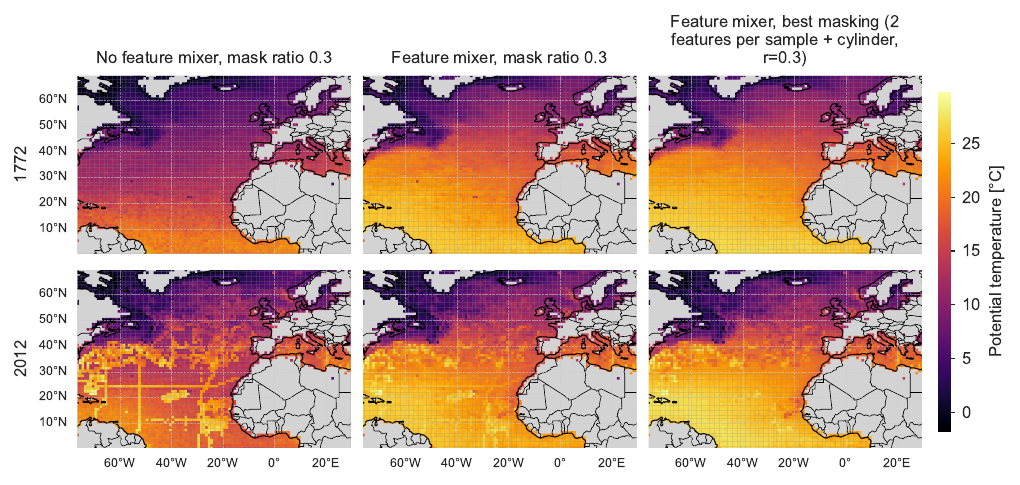}
	\caption{Imputed potential temperature fields produced by time-space attention without a feature mixer (mask ratio of 0.3), with a feature mixer (on features and missingness masks and a mask ratio of 0.3) and with a feature mixer (best masking configuration from the ablation) at the surface level in two selected 20-year bins (1772 and 2012), averaged across five seeds. Grey indicates ocean cells without data and light grey shows land}
	\label{app:fig:feature_mixer_progression_pott}
\end{figure}

\section{Uncertainties}

\begin{table}[!htbp]
	\centering
	\caption{Standardised residuals (z-scores) per feature (mean $\pm$ standard deviation across five seeds), computed as $z = (y - \hat{y}) / \hat{\sigma}_\text{total}$ from the observed value $y$, the predicted mean $\hat{y}$ and the predicted total uncertainty $\hat{\sigma}_\text{total}$ (all in $[0,1]$-scaled units), over all observed grid cells. The last column gives the number of observed values per feature}
	\label{app:tab:residuals}
	\begin{tabular}{lrrr}
		\toprule
		Feature & Mean of z & Std of z & Number of observations \\
		\midrule
		Potential temperature & 0.027 $\pm$ 0.113 & 0.887 $\pm$ 0.021 & 143,933 \\
		Salinity & $-$0.043 $\pm$ 0.163 & 0.829 $\pm$ 0.024 & 159,907 \\
		Oxygen & 0.058 $\pm$ 0.132 & 0.929 $\pm$ 0.049 & 131,326 \\
		Nitrate & 0.093 $\pm$ 0.190 & 0.888 $\pm$ 0.056 & 55,349 \\
		Silicate & 0.062 $\pm$ 0.164 & 0.937 $\pm$ 0.068 & 69,043 \\
		Phosphate & $-$0.099 $\pm$ 0.297 & 0.939 $\pm$ 0.063 & 81,744 \\
		\bottomrule
	\end{tabular}
\end{table}

\begin{figure}[!htbp]
	\centering
	\includegraphics[width=0.6\textwidth]{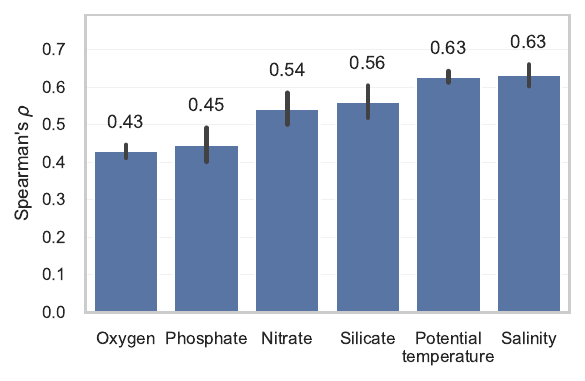}
	\caption{Spearman's $\rho$ between the total uncertainty  and the squared error per feature in $[0,1]$-scaled units, computed over all observed grid cells. The bars show the mean across five seeds and the whiskers $\pm$ one standard deviation across seeds}
	\label{app:fig:uncertainty_error_corr}
\end{figure}

\begin{figure}[!htbp]
	\centering
	\includegraphics[width=0.85\textwidth]{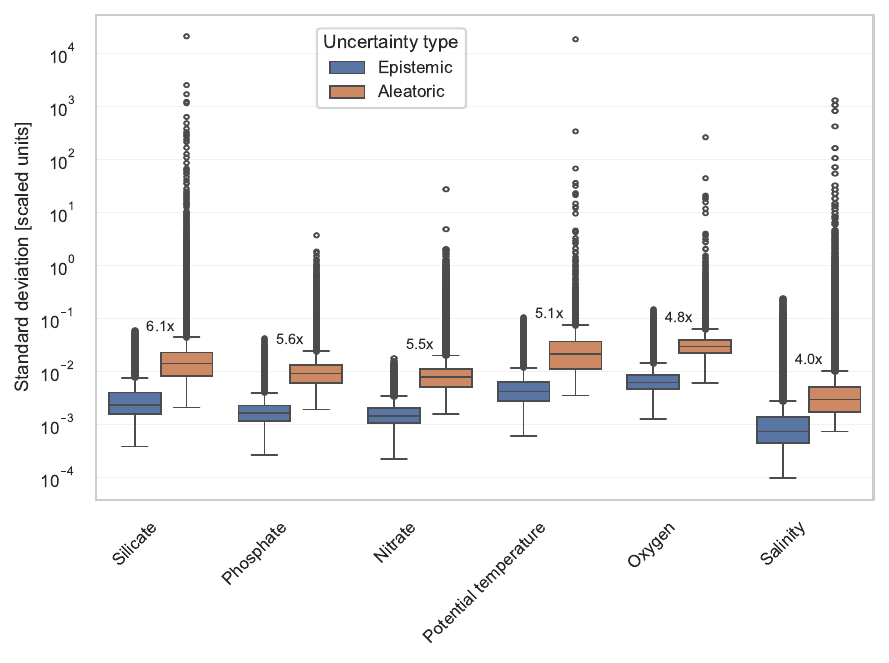}
	\caption{Boxplot of epistemic and aleatoric uncertainty distributions (standard deviation in $[0,1]$-scaled units, log scale) per feature, pooled across all \num{637390} grid cells, including cells without any observation, and five seeds. Whiskers extend to values within 1.5 times the interquartile range and circles mark values beyond. The number above each feature is the factor by which the aleatoric median exceeds the epistemic median, computed on the non-log-transformed values}
	\label{app:fig:uncertainty_per_var}
\end{figure}

\begin{figure}[!htbp]
	\centering
	\includegraphics[width=0.5\textwidth]{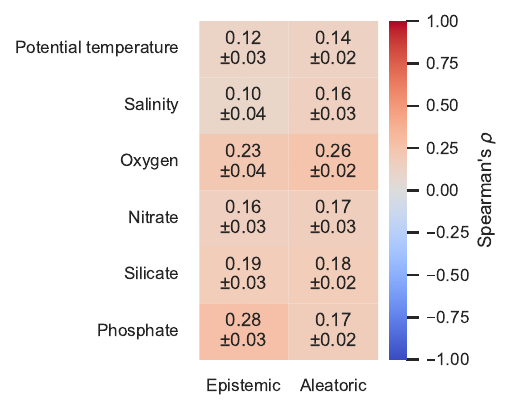}
	\caption{Spearman's $\rho$ between the epistemic and aleatoric uncertainty and the missingness of a grid cell per feature, computed over all \num{637390} grid cells. Missingness is the fraction of the six parameters without an observation in that cell. Cell values are the mean $\pm$ standard deviation across five seeds}
	\label{app:fig:uncertainty_missingness_corr}
\end{figure}

\end{appendix}

\end{document}